\documentclass{article}
\usepackage{float}
\usepackage{placeins}
\usepackage{silence}
\PassOptionsToPackage{numbers,sort}{natbib}
\usepackage[eandd, final]{neurips_2026}

\usepackage[utf8]{inputenc}
\usepackage[T1]{fontenc}
\usepackage{courier}
\usepackage{mathptmx}
\usepackage[pdfencoding=auto,colorlinks=true,linkcolor=blue,citecolor=blue,urlcolor=blue,pdftitle={When Are Predictions Enough? An Evaluation Protocol for Frozen Expert Composition},pdfauthor={Alessandro Pereira, Keith J Ransom, Lewis Mitchell}]{hyperref}
\usepackage{url}
\usepackage{booktabs}
\usepackage{amsfonts}
\usepackage{amsmath}
\usepackage{nicefrac}
\usepackage{microtype}
\usepackage{xcolor}
\usepackage{graphicx}
\usepackage{multirow}
\usepackage{array}

\newcommand{\anonrepo}{\url{https://github.com/rthinker/frozen-expert-eval}}
\newcommand{\datasethost}{\url{https://huggingface.co/datasets/anon-frozen-experts/df40-frozen-experts-anon}}
\newcommand{\psmethod}[1]{\ifmmode\mathrm{PS}_{\mathrm{#1}}\else\textup{PS\textsubscript{#1}}\fi}
\newcommand{\fsmethod}[1]{\ifmmode\mathrm{FS}_{\mathrm{#1}}\else\textup{FS\textsubscript{#1}}\fi}
\newcommand{\app}[1]{\hyperref[app:#1]{Appendix~#1}}

\title{When Are Predictions Enough? An Evaluation Protocol for Frozen Expert Composition}

\author{%
  Alessandro Pereira$^{1,4}$ \qquad Keith J Ransom$^{2}$ \qquad Lewis Mitchell$^{1,3}$ \\
  $^{1}$Australian Institute for Machine Learning (AIML), Adelaide University \\
  $^{2}$School of Computer Science and Information Technology, Adelaide University \\
  $^{3}$Adelaide Data Science Centre (ADSC), School of Mathematical Sciences, Adelaide University \\
  $^{4}$deetech.ai \\
  \texttt{\{a1759471, keith.ransom, lewis.mitchell\}@adelaide.edu.au}
}

\begin{document}

\maketitle

\begin{abstract}
Practitioners often combine independently trained models while keeping their weights frozen. These experts are commonly combined through their final predictions, although their internal features may retain useful information that those predictions omit. We ask when combining features improves on combining predictions, and how to make that comparison fair. We introduce a reusable evaluation protocol, apply it to 21 frozen deepfake detectors from DF40~\cite{yan2024}, and audit how evaluation choices change the conclusions. The protocol uses strong prediction baselines, compares simple and larger feature combiners, varies architecture and training data separately, and reports results by held-out generator with train/test overlap controls. Under leave-one-generator-out evaluation, logistic regression on the features of six architecture-diverse experts improves macro AUROC by 0.081 over the best tested prediction combiner in each fold (approximate 95\% fold-level bootstrap CI $[0.024, 0.140]$), with gains on seven of eight generators. The larger network shows no clear improvement over the linear model (approximately 3.5M versus 10K parameters), suggesting that access to richer information matters more than additional combiner capacity on this pool. Features may help when they retain information that the predictions omit and that transfers to new generators. Predictions may be enough when the strongest prediction baseline already performs well, while neither approach reliably compensates for poor expert transfer. Weakening prediction baselines on the overlap-affected standard split inflates the apparent feature advantage by up to 0.172 AUROC. Overlap-aware evaluation also changes the estimated gain, though the contrast does not isolate overlap. Results by generator and expert pool show where the advantage holds, weakens, or reverses. Without these controls, the same benchmark can support different conclusions.
\end{abstract}
\section{Introduction}

Modern backbones, foundation models, fine-tuned specialists, and off-the-shelf classifiers are widely available across computer vision~\cite{dosovitskiy2020,he2016,radford2021} and natural language processing~\cite{devlin2019}; it is routine to combine several independently trained models on a single downstream task as a frozen-expert pool.

We compare prediction-space (PS) combination, which uses experts' outputs, with feature-space (FS) combination, which uses their internal representations. Given a pool of frozen models, a common practice is to aggregate their outputs: voting or averaging predictions \cite{dietterich2000,kittler1998} or stacking predictions through a meta-learner \cite{wolpert1992}. These approaches treat each expert as a black box that emits a low-dimensional prediction, then combine those predictions into a final decision. This paradigm is simple, well-understood, and computationally inexpensive.

However, by operating on predictions alone, these methods access only a narrow projection of each expert's internal representation. A binary classifier, for instance, compresses its features into two logits before combination. When experts are heterogeneous (trained on different data distributions, built on different architectures, or both) their intermediate representations may encode substantially different aspects of the input. Prediction-space assembly can discard this structure, exposing only a low-dimensional summary of each expert's decision rather than a representation of \textit{what} the expert detected or \textit{how} its features differ from those of other experts. In practice, experts are often frozen by necessity (retraining is costly, checkpoints may be access-restricted, and joint fine-tuning may be operationally impractical), so the question of how to combine them at inference is unavoidable.

\begin{figure}[t]
\centering
\includegraphics[width=\textwidth]{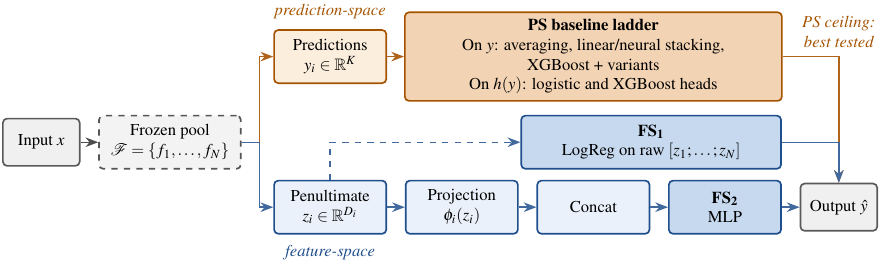}
\caption{Prediction-space (top) vs feature-space (bottom) assembly. PS combines few-class outputs $y_i \in \mathbb{R}^K$, or functions $h(y)$ of them, with a baseline ladder (\S\ref{sec:psladder}); FS combines penultimate representations $z_i \in \mathbb{R}^{D_i}$, either directly (\fsmethod{1}) or via a learned projection and MLP (\fsmethod{2}).}
\label{fig:arch}
\end{figure}

This raises a direct question: when does feature-space combination improve on prediction-space combination, and by how much? Even on the same benchmark, weak prediction baselines can inflate the apparent feature advantage, train/test overlap can distort comparisons, aggregate metrics can obscure variation across generators, and a single expert pool can give a misleading impression of generality (\S\ref{sec:sensitivity}).

Our primary contribution is a four-part evaluation protocol and accompanying sensitivity audit for assessing the size, consistency and scope of feature-space gains, and examining whether interface access or additional combiner capacity explains them.

\paragraph{Research questions.} We organize the investigation around four research questions: three empirical (Effect, Mechanism, Scope) and one targeting the evaluation methodology directly (Evaluation design):

\begin{itemize}
\item \textbf{RQ1 (Effect):} Does feature-space assembly outperform a strong prediction-space baseline ladder for frozen heterogeneous experts?
\item \textbf{RQ2 (Mechanism):} Is the gain explained by interface access rather than combiner capacity?
\item \textbf{RQ3 (Scope):} Under what pool conditions does the feature advantage persist or weaken, including under an overlap control?
\item \textbf{RQ4 (Evaluation design):} How does protocol design (baseline strength, overlap control, reporting granularity, pool construction) change supported interpretation on the same data?
\end{itemize}

We investigate these questions in the regime where prediction interfaces are low-dimensional: frozen heterogeneous experts with binary or few-class outputs ($K\!=\!2$ in our main study; a three-class check appears in \app{I}). Rather than proposing a new combiner, we compare two composition interfaces (prediction-space and feature-space; defined formally in \S\ref{sec:setting}) under a disciplined ladder of baselines, with all experts frozen throughout and only the combiner trained.

Our testbed is deepfake detection, based on the DF40 benchmark and release~\cite{yan2024}, which provides 40 generation methods together with public code and detector checkpoints. From these artifacts we construct a 21-detector frozen-expert testbed over the training variants and architectures interface-compatible with our protocol. We use DF40 as a structured instantiation of the binary-output regime. The DF40 split files place many held-out test images in the training aggregate, so the leave-one-generator-out (LOGO) control gives our primary estimates (\S\ref{sec:setup}, \app{A}). This setting lets us build pools that vary one heterogeneity axis while holding the other fixed: data diversity (same architecture, different training) or architecture diversity (same training, different architecture).

A fair assessment of whether predictions are enough must account for baseline strength, train/test overlap, reporting by held-out group, and expert-pool construction. We address RQ1 in \S\ref{sec:rq1}, RQ2 in \S\ref{sec:mechanism}, RQ3 in \S\ref{sec:scope}, and RQ4 in \S\ref{sec:sensitivity}.

\paragraph{Contributions.}
\begin{enumerate}
\item \textbf{Evaluation-design sensitivity audit} that shows, on a single benchmark, how four evaluation choices (baseline strength, overlap control, reporting granularity, pool construction) change the size, consistency, and scope of the apparent feature advantage.
\item \textbf{Reusable four-component protocol} (prediction-space baseline ladder, deterministic feature-space anchor with capacity check, axis-isolated pools, grouped evaluation with overlap-aware controls), documented as a standalone Evaluation Protocol Card (\app{J}).
\item \textbf{Evidence on mechanism and scope}: capacity and input-width controls examine explanations for the feature advantage, while comparisons across expert pools identify settings where the average advantage weakens or reverses.
\item \textbf{DF40 instantiation} on 21 frozen deepfake detectors: on the architecture-diverse pool, the linear feature combiner outperforms the strongest tested prediction baseline by \textbf{+0.081} AUROC under LOGO (95\% CI $[+0.024, +0.140]$); the standard split, which reuses test images in training, gives +0.105.
\end{enumerate}

\section{Related Work}\label{sec:related-work}

\paragraph{Prediction-space combination.} Classical ensemble work shows that combining diverse predictors can improve generalization~\cite{breiman1996,dietterich2000}: stacking~\cite{wolpert1992} trains a meta-learner on base-model predictions, and bagging~\cite{breiman1996} constructs diverse base learners during training. The classifier-combination literature~\cite{kittler1998} studies decision-level combiners under probabilistic assumptions, including classifiers with different feature representations. It combines class outputs rather than comparing prediction- and feature-space access to the same frozen expert pool. Deep ensembles~\cite{lakshminarayanan2017} use prediction-space averaging as an uncertainty baseline. Knowledge distillation~\cite{hinton2015} compresses teacher output distributions or logits into a single student, whereas we retain all experts at inference time and ask which composition interface best exposes their cross-expert information. Our work asks whether prediction-space composition is adequate when base models are frozen and heterogeneous, a setting where the practitioner inherits pre-trained experts rather than constructing them for diversity.

\paragraph{MoE and parameter-space fusion.} Mixture-of-experts (MoE) systems combine expert outputs through a learned gate; sparse variants route inputs to selected subsets~\cite{jacobs1991,shazeer2017,fedus2022}, with implementations like Switch Transformer~\cite{fedus2022} and GShard~\cite{lepikhin2020} training experts jointly under a shared architecture. Our setting differs in two respects: experts are frozen rather than jointly trained, and have different backbones and feature dimensionalities; we do not route inputs but combine outputs across all experts. A separate line of work shows that averaging or manipulating weights of models fine-tuned from a common initialization can yield strong performance without additional training~\cite{wortsman2022,ilharco2023,yadav2023}. These methods operate in parameter space and assume architectural compatibility (typically a shared pre-trained checkpoint), making them inapplicable to our heterogeneous experts (e.g., CLIP-ViT vs Xception). Feature-space assembly provides an alternative that accommodates arbitrary architectures through concatenation or per-expert projection.

\paragraph{Frozen feature/model composition.} ComBo~\cite{ramtoula2025} probes and combines features from multiple frozen backbones, comparing multi- against single-backbone probing rather than feature-space versus prediction-space composition. MoFE~\cite{seo2025} builds a shared-architecture MoE LLM with frozen FFN expert blocks and trainable shared/router parameters; despite the name, it is a parameter-efficient fine-tuning method rather than a protocol for composing independently trained heterogeneous classifiers. ZipIt!~\cite{stoica2024} merges same-architecture task-specific models with different initializations via activation-correlation feature alignment and weight fusion; it does not address heterogeneous backbones or feature dimensionalities. FusionBench~\cite{tang2025} benchmarks model ensemble, merging, and mixing across tasks and fine-tuning regimes. Closest to our domain, E3~\cite{azizpour2024e3} fine-tunes generator-specific detectors, freezes their embedders and trains an embedding-fusion network; an ablation reports gains over majority voting. Our setting uses inherited frozen third-party experts and compares interfaces against a strong prediction-space ladder while auditing evaluation choices. Each differs from our setting in at least one of three ways: shared architecture or parameter compatibility assumed, no explicit prediction-space-vs-feature-space comparison on the same frozen pool, or no overlap-aware controls and reusable evaluation card.

\paragraph{Deepfake forensics.} DF40~\cite{yan2024} provides a standardized pool of detectors trained on different forgery types; we use it as a controlled testbed for the composition-interface question, exploiting its structured variation in training data and architecture. Recent work on cross-dataset deepfake generalization~\cite{yermakov2026} studies non-compositional, parameter-efficient adaptation of mostly frozen vision encoders across 14 deepfake benchmarks, complementary to our composition focus. Prior work has studied transfer across forgery domains~\cite{cozzolino2018} and fairness across demographic groups~\cite{ju2024}. These studies address different questions from prediction-versus-feature composition on the same frozen expert pool.

\paragraph{Evaluation methodology.} Holistic benchmarking~\cite{liang2023helm}, distribution-shift benchmarks~\cite{koh2021wilds}, and domain-generalization evaluation~\cite{gulrajani2020domainbed} share the framing that evaluation choices can directly impact conclusions. We instantiate this framing for frozen-expert composition: a component-wise protocol with an explicit design-sensitivity audit (Table~\ref{tab:audit}) that shows how four evaluation choices change conclusions on the same benchmark. The protocol's components can be reused for frozen-expert pools with few-class outputs and accessible penultimate features.

\section{Evaluation Protocol}\label{sec:protocol}

We propose a four-part protocol for assessing apparent performance gains when composing frozen heterogeneous expert models, including:

\begin{itemize}
\item[\textbf{(i)}] \emph{baseline strength}: a strong prediction-space baseline ladder (\S\ref{sec:psladder});
\item[\textbf{(ii)}] \emph{combiner capacity}: a deterministic feature-space anchor plus a higher-capacity check (\S\ref{sec:fs});
\item[\textbf{(iii)}] \emph{pool construction}: axis-isolated pools (\S\ref{sec:pools});
\item[\textbf{(iv)}] \emph{overlap control and reporting granularity}: grouped evaluation controls, including overlap-aware checks and paired group-level uncertainty (\S\ref{sec:controls}).
\end{itemize}
Components (i), (iii), and (iv) address the four audit axes examined in \S\ref{sec:sensitivity}; component (ii) is orthogonal to the audit and addresses the mechanism question of whether feature-space gains arise from interface access or from combiner capacity (RQ2). This section defines the reusable protocol; Section~\ref{sec:instantiation} gives the DF40 instantiation, and \app{J} provides the protocol as a compact Evaluation Protocol Card following the Data/Model Card pattern~\cite{gebru2021datasheets,mitchell2019modelcards}.

\subsection{Comparison Setting}\label{sec:setting}

Each frozen expert exposes both a low-dimensional prediction and a higher-dimensional penultimate representation. Formally, let each frozen expert return $f_i(x) = (z_i, y_i)$, where $z_i \in \mathbb{R}^{D_i}$ is a penultimate representation and $y_i \in \mathbb{R}^K$ its prediction. Experts are heterogeneous in architecture ($D_i$ may differ) and training distribution. A combiner $g$ produces $\hat{y}$ with only $g$'s parameters trained: in \textbf{prediction-space assembly (PS)}, $g$ sees only $(y_1, \ldots, y_N)$; in \textbf{feature-space assembly (FS)}, $g$ sees only $(z_1, \ldots, z_N)$, as illustrated in Figure~\ref{fig:arch}. Because $y_i$ is a function of $z_i$ via the expert's frozen head, FS access exposes at least as much information as PS and may retain additional task-relevant information. Whether a trained combiner can exploit any such additional information under held-out-distribution evaluation is an empirical question.

\subsection{Prediction-Space Baseline Ladder}\label{sec:psladder}

To avoid overstating feature advantages through weak prediction baselines, the protocol requires a ladder spanning simple aggregation, linear stacking, neural stacking, and a tuned nonlinear combiner. Candidates may use the experts' predictions directly or derived functions of those predictions. We include cross-validation-tuned gradient-boosted trees~\cite{chen2016xgb} alongside simpler methods; no family is assumed to provide the strongest comparator across evaluation settings.

Each candidate is fitted and tuned using training and validation data only. For reporting, we define the \emph{PS ceiling} as the candidate that performs best on the test data: by macro AUROC on the standard split, or separately within each held-out-generator fold under LOGO. This retrospective choice favors prediction-based methods~\cite{cawley2010}: it narrows observed feature gains and widens feature losses (\app{F}). The ceiling is the best tested comparator, not an upper bound on prediction-based performance.

Reported confidence intervals condition on the selected comparators; they do not provide post-selection inference over the full method family. Deployment decisions should instead select the prediction-based combiner using training and validation data only.

The DF40 implementation uses 20 methods for the two primary pools on the standard split and 22 candidates per fold for the four LOGO pools (\S\ref{sec:instantiation}). The first four are probability averaging (\psmethod{1}), logistic stacking on logits (\psmethod{2}), MLP stacking on logits (\psmethod{3}), and XGBoost on logits (\psmethod{4}). Other standard-split pools use narrower subsets, so their comparisons do not establish superiority over the full ladder. Appendices~\hyperref[app:A]{A}, \hyperref[app:B]{B} and~\hyperref[app:F]{F} give the candidate sets, input transformations and tuning settings.

\subsection{Feature-Space Anchor and Capacity Check}\label{sec:fs}

We use two feature-space comparators: a simple, deterministic linear anchor (\fsmethod{1}) and a generic higher-capacity neural combiner (\fsmethod{2}). \fsmethod{1} applies L2-regularized logistic regression directly to the concatenated raw features, $[z_1; \ldots; z_N]$ of dimension $\sum_i D_i$, as practitioners receive them: without feature standardization or learned projections. Standardization and learned-projection controls (\app{B}) assess its sensitivity to raw feature scaling.

\fsmethod{2} maps each expert's features to a common dimension $d$ through a learned projection $\phi_i$, concatenates the results into an $N \cdot d$-dimensional vector, and applies an MLP combiner, where $\phi_i(z_i) = \text{ReLU}(\text{LayerNorm}(W_i z_i + b_i))$.

Comparing \fsmethod{1} with the PS ladder tests whether feature access helps even with a simple linear combiner. Comparing \fsmethod{2} with \fsmethod{1} is then a supporting capacity check: it tests whether learned projections and nonlinear combination provide further gains, although the two methods also differ in regularization, checkpoint selection, and seed aggregation.

\subsection{Axis-Isolated Pool Construction}\label{sec:pools}

Axis-isolated pools vary one heterogeneity axis at a time while holding the other fixed, and the protocol requires such pools before interpreting FS-PS deltas; boundary pools such as weak-expert or low-heterogeneity settings are used to map where the conclusion weakens. This prevents a single convenience pool from determining the apparent generality of the interface claim. Pools also differ in size, expert strength, and representation width, so comparisons across pools remain descriptive.

\subsection{Grouped Evaluation Controls}\label{sec:controls}

The protocol reports per-group performance and paired uncertainty at the held-out-group level, controls overlap between combiner training/selection and test data, and separately documents frozen experts' prior exposure to test data. PS and FS combiners use identical TRAIN/SELECT/TEST partitions and seed schedules. The resampling or fold unit should match the natural held-out group. When combiner training or selection overlaps with test data, the overlap-aware estimate (here leave-one-group-out) should lead, with the standard split reported as a secondary, overlap-affected estimate rather than as separate confirmation.

\section{Experiment}\label{sec:instantiation}

We instantiate the protocol on the DF40 deepfake detection benchmark, specifying the frozen-expert inventory, pools, splits, training settings, metrics, and uncertainty before the results are presented.

\subsection{DF40 Frozen-Expert Testbed}\label{sec:setup}

DF40~\cite{yan2024} covers 40 generation methods across four forgery families: face swapping, face reenactment, entire-face synthesis, and face editing. Its release provides public code and checkpoints through a workflow built on DeepfakeBench~\cite{yan2023deepfakebench}.

Our testbed comprises 21 frozen binary detectors, all available combinations of six detector/backbone variants\footnote{CLIP-base and CLIP-large~\cite{radford2021}, Xception~\cite{chollet2017xception}, SRM~\cite{luo2021srm}, RECCE~\cite{cao2022recce}, and RFM~\cite{wang2021rfm}.} and four training variants (face swapping, face reenactment, entire-face synthesis, and the combined DF40 training aggregate). The aggregate is a pooled training variant rather than an additional forgery family. Face-editing-trained detectors and the e4e data subset are excluded for the reproducibility reasons documented in \app{A}.

We evaluate on the test sets of eight held-out generators, DF40's unknown domains other than e4e. Held out does not mean unseen: many of their test images also appear in DF40's training aggregate, and so in the standard split's TRAIN or SELECT and in the training of aggregate-trained experts; LOGO removes each generator from combiner training and selection only. Individual experts' mean AUROC across these generators ranges from 0.36 to 0.88 (\app{H}, which lists all 21 detectors). \app{A} documents train/test overlap, and \app{F} details the LOGO control.

\subsection{Pools, Splits, and Training}

\paragraph{Primary pools.}

We construct two axis-isolated primary pools. \emph{clip4} combines the four CLIP-large experts, one per training variant, varying training data while holding architecture fixed. \emph{arch6} combines the six detector/backbone variants trained on the face-swap partition, varying architecture while holding training data fixed. \app{G} describes the additional boundary and mixed-axis pools.

\paragraph{Splits.}

For the standard split, we deduplicate \texttt{DF40\_all} by sample identifier and divide it into TRAIN and SELECT, stratified by binary label. Both come from DF40's training partitions and so share source distribution with some frozen-expert training data. Many test images of the eight held-out generators also occur in \texttt{DF40\_all}, making the standard split overlap-affected (\app{A}).

Under LOGO, we hold out one generator's test set and pool the other seven test sets as the source of combiner training data. We exclude rows whose cached features are identical to those of a held-out test image, then divide the remaining rows into TRAIN and SELECT. Each combiner is evaluated on the held-out generator, yielding eight folds. LOGO provides our primary estimates; the standard split is secondary. \app{F} details the folds and exclusions, and \app{A} examines repeated content within TRAIN and between TRAIN and SELECT in both protocols.

\paragraph{Training.}

Combiners are fitted on TRAIN, with calibration, static weights, and model selection on SELECT or by cross-validation within TRAIN, as applicable. Stochastic methods are trained with eight seeds, with predictions averaged before calculating AUROC, so their results describe eight-model ensembles (\app{C} reports seed variation for the configurations it checks); deterministic methods are not. \app{A} gives the full training settings.

\subsection{Metrics and Uncertainty}\label{sec:metrics}

Our primary metric is macro AUROC: the mean AUROC across the eight held-out generators, giving each generator equal weight. We calculate the FS$-$PS difference for each generator and bootstrap these paired differences to obtain a 95\% confidence interval for their mean~\cite{efron1979}. Under LOGO, each generator corresponds to one test fold.

We also report fold win counts and two-sided sign, Wilcoxon signed-rank, and exact sign-flip tests (\app{F}). Because LOGO folds share training data, the intervals and $p$-values are approximate and are not adjusted for multiple comparisons. The nominal 95\% percentile interval also undercovers in the eight-group simulation of \app{F}; its exclusion of zero should be read alongside the paired tests. All differences are calculated before rounding.

\section{Results}\label{sec:results}

\subsection{RQ1: Feature Access vs Prediction Access}\label{sec:rq1}

\textit{Answer.} Under LOGO, \fsmethod{1} achieves higher macro AUROC than the PS ceiling on the architecture-diverse pool (arch6), but lower macro AUROC on the same-architecture pool (clip4). For clip4, the feature advantage appears only on the overlap-affected standard split.

\paragraph{Architecture-diverse pool (arch6).}

Under LOGO, \fsmethod{1} achieves a macro AUROC of 0.898 versus 0.817 for the PS ceiling, a gain of \textbf{+0.081} (fold-level bootstrap 95\% CI $[+0.024, +0.140]$), with gains on seven of eight generators, one of them negligible (HeyGen, +0.001 on 101 test samples; Table~\ref{tab:main}). Although the bootstrap interval excludes zero, none of the three paired tests reaches $p\!<\!0.05$ (exact sign-flip: $p\!=\!0.055$; \app{F}).

On the overlap-affected standard split, the gain is larger: +0.105 (95\% CI $[+0.058, +0.158]$; \app{B}). A strict path-disjoint refit of the standard split (no training rows whose paths occur in a test set; a narrower prediction ladder) reverses the margin to $-0.228$. However, it also removes most real training images and substantially reduces their source diversity, so it cannot isolate the effect of overlap (\app{F}).

\begin{table}[ht]
\caption{\fsmethod{1} versus the PS ceiling, measured by macro AUROC across the eight held-out generators. Under LOGO, the PS ceiling is selected retrospectively on each fold's test data, favoring PS. $\Delta$ = \fsmethod{1} $-$ PS ceiling; wins count generators with $\Delta\!>\!0$.}
\label{tab:main}
\centering
{\footnotesize
\begin{tabular}{llrrrr}
\toprule
\textbf{Pool} & \textbf{Protocol} & \multicolumn{1}{l}{\textbf{PS ceiling}} & \multicolumn{1}{l}{\textbf{\fsmethod{1}}} & \multicolumn{1}{l}{\textbf{$\Delta$}} & \multicolumn{1}{l}{\textbf{Wins}} \\
\midrule
arch6 & LOGO & 0.817 & 0.898 & +0.081 & 7/8 \\
clip4 & LOGO & 0.921 & 0.846 & $-$0.074 & 5/8 \\
strong4 & LOGO & 0.910 & 0.850 & $-$0.060 & 5/8 \\
weak6 & LOGO & 0.721 & 0.619 & $-$0.102 & 3/8 \\
\midrule
arch6 & Standard & 0.860 & 0.966 & +0.105 & 8/8 \\
\bottomrule
\end{tabular}
}
\\[2pt]
{\scriptsize\itshape The standard split is overlap-affected. Differences are calculated before rounding. strong4 comprises four high-scoring CLIP-based experts and weak6 the six weakest, both chosen using individual macro AUROC on the test generators (\app{G}).}
\end{table}

\paragraph{Same-architecture pool (clip4).}

Under LOGO, \fsmethod{1} achieves a macro AUROC of 0.846 versus 0.921 for the PS ceiling, a difference of $-0.074$ despite gains on five of eight generators (Table~\ref{tab:main}). Section~\ref{sec:scope} examines how these losses are distributed across generators.

On the overlap-affected standard split, the ordering reverses: \fsmethod{1} exceeds the PS ceiling by +0.024 (95\% CI $[+0.007, +0.043]$), with gains on all eight generators (\app{B}).

\subsection{RQ2: Interface Access vs Capacity}\label{sec:mechanism}

\textit{Answer.} On arch6, the feature advantage is obtained with a simple linear combiner, with little additional benefit from a much larger network. Width controls further support the role of richer information on the standard split; under LOGO, 12 label-fitted values still exceed \psmethod{4} but stay slightly below full-feature \fsmethod{1}.

\paragraph{Capacity checks.} Moving from the 10K-parameter \fsmethod{1} to the 3.5M-parameter \fsmethod{2} adds +0.004 AUROC on the arch6 standard split and gives no clear improvement under LOGO (\fsmethod{2} $-$ \fsmethod{1}: $-0.002$, 95\% CI $[-0.018, +0.012]$). This pattern is pool-dependent: under LOGO, \fsmethod{2} has higher point estimates than \fsmethod{1} on clip4, strong4 and weak6 (by +0.046, +0.031 and +0.014), while still trailing their PS ceilings ($-0.028$, $-0.028$ and $-0.089$). Comparing the same combiner family across interfaces also favors features: XGBoost on features exceeds \psmethod{4} by +0.142 on arch6 under LOGO, with gains on all eight generators (Table~\ref{tab:rq2}).

\paragraph{Width controls.} On arch6, features supply 9,984 values per sample, compared with 12 prediction values. We reduce each expert's features to two values and fit \fsmethod{1} on the resulting 12 inputs. Both mappings are fitted on TRAIN: label-fitted values separate real from fake, while PCA values capture variation without using labels.

On the overlap-affected standard split, label-fitted values approach full-feature \fsmethod{1} performance (0.958 versus 0.966), whereas PCA values give a lower point estimate than \psmethod{4} (0.740 versus 0.843; sign-flip $p\!=\!0.109$). Under LOGO, where \psmethod{4} (0.727) sits well below the PS ceiling (0.817), PCA values achieve 0.730, close to \psmethod{4} and below the ceiling, while label-fitted values exceed both (0.880) but remain below full-feature \fsmethod{1} (difference $-0.018$, 95\% CI $[-0.043, -0.001]$). \app{D} reports the full comparisons and the bottleneck analysis, and \app{A} examines sensitivity to duplicated training rows.

These controls suggest that the gain depends on which information reaches the combiner, but do not establish causality. Matching input width does not equalize total learned capacity: constructing the label-fitted values still uses the full features and training labels.

\subsection{RQ3: Scope and Boundary Conditions}\label{sec:scope}

\textit{Answer.} Among the four pools tested under LOGO, only arch6 shows a positive average \fsmethod{1}$-$PS difference. The high-ceiling pools, clip4 and strong4, favor predictions on average, while weak6 shows limited performance through either interface.

\paragraph{Scope.} In clip4 and strong4, \fsmethod{1} trails the PS ceiling by 0.074 and 0.060, respectively (Table~\ref{tab:main}). Much of the negative average in each pool comes from HeyGen and DeepFaceLab, the two smallest test sets. \app{F} reports an exploratory analysis excluding these generators; the primary comparison retains all eight. In weak6, the PS ceiling reaches 0.721 AUROC and \fsmethod{1} reaches 0.619, indicating limited transfer through both interfaces. Feature access therefore does not consistently compensate for a weak expert pool. The 95\% intervals for clip4, strong4, and weak6 are wide ($[-0.182, -0.002]$, $[-0.154, +0.007]$, and $[-0.212, -0.013]$), and none of the three paired tests reaches $p\!<\!0.05$ on any of these pools.

Both clip4 and strong4 include an expert trained on the DF40 aggregate, whose training list contains many held-out test images. LOGO removes the identified reuse in combiner training but cannot undo this expert-training exposure, which remains a limitation when interpreting these pools (\app{A}).

\subsection{RQ4: Evaluation-Design Sensitivity}\label{sec:sensitivity}

\textit{Answer.} Restricting prediction baselines inflates the apparent feature advantage by up to \textbf{+0.172} AUROC on the standard split. Overlap control, reporting granularity, and pool construction also change the size, consistency, and scope of the result.

\paragraph{Sensitivity audit.} Table~\ref{tab:audit} compares practical evaluation choices along four axes.

\paragraph{Baseline strength.} Adding \psmethod{4} on weak6's standard split sharply reduces the apparent feature advantage, showing why linear and neural prediction baselines alone are insufficient here. Under LOGO, the corresponding inflation from limiting the comparator to \psmethod{1}-\psmethod{3} is at most +0.026 across the four pools (\app{F}).

\paragraph{Pool construction.} Variation across LOGO pools limits generalization from any single pool. The standard-split pool comparison is additionally qualified by unequal PS-ladder coverage: narrower ladders yield upper bounds on the corresponding full-ladder margins (\app{G}).

\paragraph{Overlap control.} The arch6 standard--LOGO contrast changes both training data and PS-ceiling selection, so it does not isolate overlap. With the same selection rule in both protocols, the contrast is $-$0.006 or +0.011, depending on the rule (\app{F}).

\paragraph{Reporting granularity.} On weak6's standard split, the positive average \fsmethod{2} difference has uneven support across generators; the mean alone obscures that variation.

The magnitudes also depend on duplicated training rows. Removing exact duplicates changes the baseline inflation from +0.074 to +0.012 on arch6 and from +0.172 to +0.096 on weak6, and the arch6 standard--LOGO contrast from +0.025 to +0.087 (\app{A}). All three remain positive, but their sizes are sensitive to duplicate handling.

\begin{table}[H]
\caption{How evaluation choices change the reported feature advantage, and which protocol component addresses each choice. $\Delta$ denotes FS macro AUROC minus the strongest prediction baseline included in the comparison. Results use \fsmethod{1} unless otherwise stated.}
\label{tab:audit}
\centering
{\footnotesize
\setlength{\tabcolsep}{5pt}
\begin{tabular}{>{\raggedright\arraybackslash}p{2.3cm}>{\raggedright\arraybackslash}p{2.3cm}>{\raggedright\arraybackslash}p{4.4cm}>{\raggedright\arraybackslash}p{3.2cm}}
\toprule
\textbf{Evaluation choice} & \textbf{Protocol component} & \textbf{Comparison} & \textbf{Change in reported conclusion} \\
\midrule
Baseline strength & (i) Strong PS baseline ladder & Standard split: expanding \psmethod{1}-\psmethod{3} to the tested ladder changes $\Delta$ from ${+}0.179$ to ${+}0.105$ on arch6, and from ${+}0.232$ to ${+}0.060$ on weak6. & Weaker baselines inflate the apparent advantage by $\mathbf{{+}0.074}$ and $\mathbf{{+}0.172}$, respectively. \\
\midrule
Pool construction & (iii) Axis-isolated pools & $\Delta$ ranges from ${+}0.011$ to ${+}0.107$ across nine standard-split pools, and from ${-}0.102$ to ${+}0.081$ across four LOGO pools. & The magnitude, and under LOGO the sign, depends on the pool. \\
\midrule
Overlap control & (iv) Grouped evaluation controls & arch6: $\Delta$ changes from ${+}0.105$ on the standard split to ${+}0.081$ under LOGO. & A $\mathbf{{+}0.025}$ protocol difference; training data and PS selection also change, so this does not isolate overlap. \\
\midrule
Reporting granularity & (iv) Grouped evaluation controls & weak6, \fsmethod{2}, standard split: mean $\Delta = {+}0.022$, with wins on 5/8 generators and sign-test $p\!=\!0.73$. & The positive average has uneven support across generators. \\
\bottomrule
\end{tabular}
}
\\[2pt]
{\scriptsize\itshape Standard-split results are overlap-affected. Differences are calculated before rounding. Seven of the nine standard-split pools use narrower prediction ladders, so their margins are upper bounds on the corresponding full-ladder margins. \app{A} reports sensitivity to duplicated training rows.}
\end{table}

\section{Discussion}\label{sec:discussion}

\paragraph{Practical guidance: when feature access is worth the cost.} Our findings motivate five checks:
\begin{enumerate}
\item \emph{Establish a strong prediction baseline.} Compare several PS methods and select among them using TRAIN/SELECT. The retrospective PS ceiling is an evaluation reference, not a deployment selection rule.
\item \emph{Check whether the outputs match the task.} If the task requires classes the experts were not trained to predict, consider fitting task-specific heads on their frozen features before concluding that PS is insufficient (\app{I}.1).
\item \emph{Assess expert transfer.} Evaluate both interfaces on representative held-out groups. If both perform poorly, prioritize improving the expert pool before investing in a more complex combiner.
\item \emph{Check feature access and additional cost.} Confirm that the experts expose usable features, and assess the additional storage, data-transfer, training, and inference costs. Prediction-only APIs may rule out FS (\app{I}).
\item \emph{Run a held-out-group comparison.} Compare a simple feature combiner against the selected prediction baseline on held-out groups, using matched splits and overlap checks. Inspect per-group performance and uncertainty, then judge whether the observed benefit justifies the additional cost.
\end{enumerate}

\paragraph{Scope of the claims.} Our empirical claims concern the tested expert pools and evaluation settings; they do not establish that feature access is generally superior. \fsmethod{1} is a simple evaluation reference, not an established best deployment choice. The prediction ladder covers a finite set of methods and searches, and stronger baselines could narrow the observed feature gains. The width controls do not establish that input dimensionality is irrelevant. Evidence is primarily for binary outputs, with limited three-class checks (\app{I}.1); many-class and structured-output tasks remain untested. Calibration, fairness, and deployment robustness require separate evaluation.

\paragraph{Limitations.} The main evidence comes from one deepfake benchmark and eight held-out generators; the protocol's broader validity remains to be established. The small number of generators and shared training data across LOGO folds limit statistical inference, so the reported intervals and $p$-values should be treated as approximate (\app{F}).

Standard-split results are overlap-affected. LOGO removes the identified reuse in combiner training, but identical-feature checks cannot rule out re-encoded copies. Repeated content also remains within TRAIN and between TRAIN and SELECT in both protocols, potentially affecting model selection; repeated test rows additionally affect the weighting of images in row-based metrics; \app{A} reports sensitivity to duplicated training rows. Knowledge of expert-training exposure is limited to DF40's training lists (\app{A}).

Responsible use must account for misuse and the demographic and privacy limitations inherited from DF40. The feature cache's dataset card and Croissant metadata document provenance, intended uses, and these limitations (\app{J}.9).

\section{Conclusion}\label{sec:conclusion}

We introduce a reusable protocol and sensitivity audit for evaluating when feature access helps in frozen-expert composition. On DF40, a linear feature combiner achieves an observed macro AUROC gain of +0.081 over the per-fold PS ceiling on the architecture-diverse pool under LOGO. A much larger feature combiner provides no clear improvement on this pool, while the linear anchor trails the PS ceiling on the other three pools tested under LOGO. The benefit of feature access is therefore conditional on the expert pool and evaluation setting.

The audit shows how baseline strength, overlap control, reporting granularity, and pool construction affect the size, consistency, and scope of the apparent advantage. Practitioners should assess whether feature gains survive strong prediction baselines, overlap controls, and evaluation on held-out groups before deciding whether they justify the additional cost. The protocol and the Evaluation Protocol Card provide concrete checks for making this assessment; broader validation across benchmarks and tasks remains necessary. The accompanying implementation, verification harness, and derived feature cache (\app{J}.9) support reproduction and application to new expert pools.

\begin{ack}
Alessandro Pereira was supported by an Australian Government Research Training Program (RTP) Scholarship. No other funding supported this work.

Alessandro Pereira is the founder of deetech.ai, a company that commercializes synthetic-media detection technology. The company provided no funding, data, or other resources for this work. Alessandro Pereira is also the applicant for a provisional patent application that may relate to this work. The other authors declare no competing interests.

The authors used AI tools to assist with writing and programming.
\end{ack}

\clearpage

\setlength{\bibsep}{8pt plus 2pt minus 1pt}

\newpage

\appendix

\renewcommand{\thetable}{A\arabic{table}}
\renewcommand{\theHtable}{A\arabic{table}}
\setcounter{table}{0}
\renewcommand{\thefigure}{A\arabic{figure}}
\renewcommand{\theHfigure}{A\arabic{figure}}
\setcounter{figure}{0}

\subsection*{Appendix Map}

\app{A}: data, training and sensitivity checks; \app{B}: full standard-split results; \app{C}: \fsmethod{2} robustness; \app{D}: representations and width controls; \app{E}: expert removal; \app{F}: primary LOGO evidence and statistical checks; \app{G}: additional pools; \app{H}: individual experts and AUPRC; \app{I}: three-class, speech and cost checks; \app{J}: reusable protocol card and release scope.

\subsection*{A. Data, Training Details, and Sensitivity Checks}\label{app:A}

\paragraph{Data splits and seeds.} We extract logits and features from our local DF40 copy using one frame per video (\texttt{frame\_num}\,=\,1), without further subsampling. The DF40 loader uses c23 inputs at $256\times256$, no augmentation, and per-channel mean and standard deviation of 0.5. We bilinearly resize CLIP inputs to $224\times224$ and spatially average detector features when they are feature maps. We exclude e4e (29,126 training entries and its test set) and face-editing-trained detector variants because we could not reproduce their reported DF40 results in our setup. Counts below refer to cached rows, not necessarily distinct images.

The standard split uses deduplication by \texttt{sample\_id}, followed by a binary-label-stratified 80/20 split of \texttt{DF40\_all}, giving 26,591 TRAIN and 6,647 SELECT rows. SELECT fits temperatures (\psmethod{A}, \psmethod{H}), averaging weights (\psmethod{B}), and neural checkpoint and weight-decay choices. \psmethod{C} and \psmethod{4} use five-fold CV within TRAIN. The split, sklearn CV folds and LOGO bootstrap use seed 1024; the standard-split generator bootstrap uses seed 42. Both bootstraps use 10,000 resamples. Stochastic methods use seeds $\{7,10,24,42,1024,1973,3392,9721\}$; deterministic methods use one fit with fixed random states where applicable. LOGO partitions are specified in \app{F}.

\paragraph{Identifier deduplication and repeated content.} TRAIN and SELECT share no sample identifiers, but different identifiers can correspond to identical cached features. On the standard split with arch6 features, 793 SELECT rows match a TRAIN row and 1,880 TRAIN rows repeat another TRAIN row; the SELECT count is 794 with clip4 features. StarGAN-v2's 4,000 test rows represent 1,024 distinct images. Across the seven LOGO folds that include StarGAN-v2 in training, the number of repeated TRAIN rows ranges from 2,335 to 2,377, and the proportion of SELECT rows matching TRAIN ranges from 23\% to 31\%. Copies can also cross inner CV folds. Repetition changes fitting weights and can affect hyperparameter or checkpoint selection.

\paragraph{Overlap between combiner training and test data.} On the standard split, 11,886 (66.3\%) of the 17,925 cached test rows occur in TRAIN or SELECT under the same path and 12,893 (71.9\%) match cached df40-clip-large features. Reuse includes real and fake images, with fake-image reuse in five generators.

\begin{center}
Standard-split overlap by test generator.\\[4pt]
{\footnotesize
\begin{tabular}{lrrr}
\toprule
\textbf{Generator} & \textbf{Test rows} & \textbf{Same-path matches} & \textbf{Additional feature matches} \\
\midrule
WhichIsReal & 2,001 & 988 & 0 \\
CollabDiff & 2,000 & 1,993 & 7 \\
MidJourney & 1,882 & 1,882 & 0 \\
HeyGen & 101 & 50 & 0 \\
StarGAN-v2 & 4,000 & 3,000 & 1,000 \\
DeepFaceLab & 149 & 149 & 0 \\
StarGAN & 3,968 & 0 & 0 \\
StyleCLIP & 3,824 & 3,824 & 0 \\
\midrule
\textbf{Total} & \textbf{17,925} & \textbf{11,886} & \textbf{1,007} \\
\bottomrule
\end{tabular}
}
\end{center}

Additional feature matches are beyond the same-path matches. Standard-split results are therefore \textbf{overlap-affected}. LOGO excludes identified feature-identical held-out content from combiner training (\app{F}); this does not rule out re-encoded copies.

\paragraph{Frozen-expert exposure and source integrity.} Pool definitions and expert naming conventions are given in \app{G} and \app{H}. Among DF40's training lists, matching test paths occur only in the aggregate list. No expert in arch6, weak6 or low\_het4 was trained on it; each other pool contains df40-clip-large, and balanced8 and full21 contain additional aggregate-trained experts. All MidJourney test paths occur in df40-clip-large's training list; its MidJourney AUROC is 0.697. These list-based checks cannot exclude other prior exposure, which LOGO combiner exclusions cannot undo.

Seventeen cached rows refer to nine distinct zero-byte source files in our local copy. For each affected row, the extraction loader substituted the first item of the corresponding split, leaving nine cached rows with labels inconsistent with their paths. These rows remain in the cache; their effect has not been separately quantified or corrected by the duplicate check below. These preprocessing choices also qualify comparisons with the DF40 authors' results.

\paragraph{Sensitivity to duplicated training rows.} We refit after removing rows whose cached features match across all experts in a pool. LOGO retains the first copy before each fold's TRAIN/SELECT split. On the standard split, TRAIN retains its first copies and SELECT removes its own repeats and all TRAIN matches. Test rows, including their multiplicities, are unchanged. Table~\ref{tab:sensitivity} summarizes the results.

\begin{table}[htbp]
\caption{Sensitivity to removing duplicated training rows.}
\label{tab:sensitivity}
\centering
{\footnotesize
\setlength{\tabcolsep}{4pt}
\begin{tabular}{>{\raggedright\arraybackslash}p{5.6cm}ll}
\toprule
\textbf{Quantity} & \textbf{Reported results} & \textbf{Training duplicates removed} \\
\midrule
arch6 LOGO: \fsmethod{1} $-$ PS ceiling & +0.081 $[+0.024, +0.140]$, 7/8 & +0.078 $[+0.025, +0.135]$, 6/8 \\
arch6 LOGO: AUPRC, \fsmethod{1} $-$ ceiling method & +0.080 & +0.065 \\
clip4, strong4, weak6 LOGO: \fsmethod{1} $-$ PS ceiling & $-$0.074, $-$0.060, $-$0.102 & $-$0.073, $-$0.053, $-$0.095 \\
arch6 LOGO: 12 PCA values $-$ \psmethod{4} & +0.003 $[-0.026, +0.033]$ & $-$0.032 $[-0.062, +0.001]$ \\
arch6 LOGO: 12 label-fitted values $-$ \fsmethod{1} & $-$0.018 $[-0.043, -0.001]$ & $-$0.023 $[-0.056, +0.002]$ \\
arch6 standard: \fsmethod{1} $-$ PS ceiling & +0.105 $[+0.058, +0.158]$ & +0.166 $[+0.113, +0.218]$ \\
clip4 standard: \fsmethod{1} $-$ PS ceiling & +0.024 $[+0.007, +0.043]$ & +0.028 $[+0.013, +0.045]$ \\
standard--LOGO contrast, arch6 & +0.025 & +0.087 \\
standard-split baseline inflation, arch6; weak6 & +0.074; +0.172 & +0.012; +0.096 \\
\bottomrule
\end{tabular}
}
\\[2pt]
{\scriptsize Macro AUROC differences unless stated; intervals are paired fold (LOGO) or generator (standard) bootstraps, and $k/8$ counts positive folds. AUPRC uses AUROC-selected comparators. Both standard-split columns remain overlap-affected. The last two rows revisit Table~\ref{tab:audit}. The alternative standard runs use the richer-input script, whose primary standard--LOGO contrast is +0.024 versus +0.025 in the audit. weak6 baseline inflation uses \psmethod{4} versus \psmethod{1}--\psmethod{3}. Differences use unrounded inputs.}
\end{table}

The signs of the mean \fsmethod{1} margins remain unchanged, but the standard-split effects and width comparisons are sensitive. Removing duplicates also changes training composition and weights: StarGAN-v2 contributes 1,024 rather than 4,000 LOGO training rows. These shifts therefore do not isolate cleaner validation. The width-control intervals after removal include zero and do not establish equivalence (\app{D}.4).

\paragraph{Sensitivity to LOGO prediction-head settings.} Refitting all 32 folds with the standard-split richer-input heads (class-weighted logistic regression, the \psmethod{4} grid with five-fold CV, and \psmethod{K}; 20 methods) leaves all four \fsmethod{1} margins unchanged at three decimals: $+0.081$, $-0.074$, $-0.060$, $-0.102$. Only HeyGen changes at the precision of Table~\ref{tab:logo}: $\Delta_1$ becomes +0.002 rather than +0.001, and $\Delta_2$ becomes $-0.038$ rather than $-0.039$.

\paragraph{Training and aggregation.} Neural combiners use Adam, learning rate $2\!\times\!10^{-4}$, batch size 256, uniform shuffling and inverse-frequency class-weighted cross-entropy. Weight decay is zero except for \psmethod{E}/\psmethod{F}. For stochastic methods, sample probabilities are averaged across the eight seeds before calculating per-generator AUROC and its unweighted macro average. This evaluates an eight-model ensemble; per-seed AUROC variation is reported separately in \app{C}.

\begin{table}[tp]
\caption{Method specifications. Settings below describe the standard-split experiments; LOGO head differences are specified in \app{F}.}
\label{tab:methods}
\centering
{\footnotesize
\setlength{\tabcolsep}{4pt}
\begin{tabular}{l>{\raggedright\arraybackslash}p{2.6cm}>{\raggedright\arraybackslash}p{2.4cm}l>{\raggedright\arraybackslash}p{5.8cm}}
\toprule
\textbf{Method} & \textbf{Description} & \textbf{Inputs} & \textbf{Det.} & \textbf{Key settings} \\
\midrule
\multicolumn{5}{l}{\textit{Core PS baselines}} \\
\psmethod{1} & Probability averaging & Probabilities & Yes & Uniform average; no fitting \\
\psmethod{2} & Logistic regression & Logits & Yes & L-BFGS, balanced class weights, $C\!=\!1$ \\
\psmethod{3} & MLP & Logits & No & One hidden layer of width 64; dropout 0.1; 10 epochs \\
\psmethod{4} & XGBoost & Logits & Yes & Hyperparameters selected by five-fold CV within TRAIN \\
\midrule
\multicolumn{5}{l}{\textit{Additional PS baselines}} \\
\psmethod{A} & Calibrated averaging & Logits $\rightarrow$ calibrated probabilities & Yes & One temperature per expert~\cite{guo2017}, fitted by class-balanced NLL on SELECT \\
\psmethod{B} & Learned averaging weights & Probabilities & Yes & Simplex-constrained weights fitted by SELECT AUROC \\
\psmethod{C} & Regularized logistic regression & Logits & Yes & Balanced class weights; penalty and $C$ selected by five-fold CV within TRAIN \\
\psmethod{D} & MLP & Probabilities & No & \psmethod{3} architecture and training settings \\
\psmethod{E} & MLP with early stopping & Probabilities & No & \psmethod{3} architecture; weight-decay sweep; patience 10; maximum 200 epochs \\
\psmethod{F} & MLP with early stopping & Logits & No & Same selection procedure as \psmethod{E} \\
\midrule
\multicolumn{5}{l}{\textit{Richer PS inputs}} \\
\psmethod{G} & Pairwise disagreement & Logits and derived values & Yes & Logits plus all pairwise $|m_i - m_j|$; LR and XGBoost heads \\
\psmethod{H} & Uncertainty summaries & Calibrated probabilities and derived values & Yes & Per-expert probabilities, their mean, entropy summaries, and dispersion statistics; LR and XGBoost heads \\
\psmethod{I} & Quadratic interactions & Logits and derived values & Yes & Logits, $m_i^2$, and pairwise $m_i m_j$; LR and XGBoost heads \\
\psmethod{J} & Rank transformation & Logit margins & Yes & Per-expert empirical CDF fitted on TRAIN; LR and XGBoost heads \\
\psmethod{K} & CV-selected input map & \psmethod{G}-\psmethod{J} inputs & Yes & Selects the map by logistic-head CV AUROC within TRAIN; reports LR and XGBoost heads on that map \\
\midrule
\multicolumn{5}{l}{\textit{FS comparators}} \\
\fsmethod{1} & Logistic regression on raw features & Concatenated features & Yes & L-BFGS, balanced class weights, $C\!=\!1$, maximum 2,000 iterations; no feature standardization \\
\fsmethod{2} & Learned projections and MLP & Features & No & Per-expert Linear $\rightarrow$ LayerNorm $\rightarrow$ ReLU; MLP $[512, 256]$; 10 epochs; best SELECT-AUROC checkpoint \\
\midrule
\multicolumn{5}{l}{\textit{FS controls}} \\
\fsmethod{A} & Learned projections and linear classifier & Features & No & Replaces \fsmethod{2}'s MLP classifier with a linear classifier \\
\fsmethod{B} & PCA and logistic regression & Concatenated features & Yes & PCA fitted on TRAIN to $d$ total dimensions, followed by logistic regression \\
\fsmethod{C} & Random projections and MLP & Features & No & Fixed Gaussian projections followed by a trained MLP \\
\fsmethod{D} & XGBoost on raw features & Concatenated features & Yes & Fixed settings; 200 trees, maximum depth 3; no CV \\
\bottomrule
\end{tabular}
}
\\[2pt]
{\scriptsize ``Det.'' means one fit with fixed random states; other methods use eight seeds. LR = logistic regression, NLL = negative log-likelihood, CV = cross-validation. Searches and mapped inputs are specified in the text; LOGO head differences are in \app{F}.}
\end{table}

\paragraph{Prediction-space searches and mappings.} \psmethod{4} searches trees $\{50,100,200\}$, depths $\{2,3,4\}$, learning rates $\{0.05,0.1,0.3\}$, and independently row and column subsampling $\{0.8,1.0\}$. \psmethod{C} searches $C\in\{0.001,0.01,0.1,1,10,100\}$ and L1, L2 or elastic-net penalties (\texttt{l1\_ratio}=0.5); L2 uses L-BFGS and the others SAGA. Both select by mean AUROC over five stratified TRAIN folds, then refit on TRAIN. All XGBoost heads use \texttt{scale\_pos\_weight}=$n_{\mathrm{neg}}/n_{\mathrm{pos}}$ and seed 1024. \psmethod{B} fits softmax-parameterized weights by Nelder--Mead from uniform weights, for at most 2,000 iterations. \psmethod{D}/\psmethod{E} use both class probabilities; \psmethod{1}/\psmethod{B} use fake probabilities.

In Table~\ref{tab:methods}, $m_i$ is expert $i$'s fake-minus-real logit margin. \psmethod{A} fits each temperature on SELECT within $[0.01,100]$. \psmethod{H} reuses these temperatures; its summaries comprise entropy of mean probability, mean expert entropy, their difference, and probability variance, minimum, maximum, median and range. Each \psmethod{G}--\psmethod{J} map has a logistic head (training-fitted standardization, balanced weights, $C=1$) and an XGBoost head (unstandardized inputs, \psmethod{4} grid). \psmethod{K} selects the map by five-fold logistic-head CV, then reports both heads. \psmethod{J}'s empirical CDFs are fitted on all available TRAIN before inner CV, so inner validation folds contribute; TEST never does. Under LOGO, these operations are fold-local, with the head and search differences specified in \app{F}.

\paragraph{Neural selection and feature controls.} \fsmethod{2} projects each expert from $D_i$ to $d$ dimensions through Linear--LayerNorm--ReLU, concatenates the outputs and applies an MLP with hidden widths $[512,256]$, ReLU and dropout 0.1. Primary $d$ is 32 for clip4 and 256 for arch6. Expert dropout independently masks each expert per sample with probability 0.25, retains at least one, and scales retained vectors by $N/n_{\mathrm{active}}$; all experts are active at evaluation. Each seed trains for 10 epochs, selecting the highest SELECT-AUROC checkpoint.

For \psmethod{E}/\psmethod{F}, the \psmethod{3} MLP takes probabilities/logits, respectively. Per seed, weight decay $\{0,10^{-4},10^{-3},10^{-2}\}$ is searched with at most 200 epochs and SELECT-AUROC patience 10. The best decay and epoch are selected, then the model is reinitialized with the same seed and refitted for that many epochs.

The clip4 controls \fsmethod{A}/\fsmethod{C} use $d=64$ and retain \fsmethod{2}'s dropout, schedule and checkpoint rule. \fsmethod{C} fixes seed-drawn $\mathcal N(0,1/D_i)$ projections but trains LayerNorm and the classifier. \fsmethod{B} uses 64 PCA components and \fsmethod{1}'s logistic settings. \fsmethod{D} uses 200 depth-3 trees, learning rate 0.1, row subsampling 0.8, column subsampling 0.5, \texttt{hist} and seed 1024, with class balancing and no CV. It matches \psmethod{4}'s model family, not its tuning procedure.

\paragraph{Computation.} The main DF40 runs use an RTX 5080 (16\,GB). The approximate 2\,h estimate for GPU training and bootstraps on cached inputs does not specify which splits it covers. On a 16\,GiB Intel i9-10980HK laptop, the camera-ready harness is estimated at 36\,h from separately timed parts, with roughly 15\,h more for the separate sensitivity reruns. CPU-bound \psmethod{E}/\psmethod{F} sweeps dominate wall-clock. This excludes upstream feature extraction, speech and t-SNE; the full21 \fsmethod{1} fit is skipped on that host for memory. Execution coverage, tolerances and retained outputs are documented in \app{J}.9 and the release.

\subsection*{B. Full PS Ladder and FS Results}\label{app:B}

These standard-split results are overlap-affected (\app{A}); Table~\ref{tab:main} and \app{F} give the primary LOGO comparisons. Table~\ref{tab:ladder} includes the four richer prediction maps and CV-selected map of \app{A}, alongside the original ladder and feature controls.

\begin{table}[htbp]
\caption{Full prediction-space baseline ladder and feature-space results on the standard split (overlap-affected). Macro AUROC is the unweighted mean across the eight generators. Bold identifies the PS ceiling in each pool.}
\label{tab:ladder}
\centering
{\footnotesize
\begin{tabular}{lrrrr}
\toprule
 & \multicolumn{2}{c}{\textbf{clip4 (4 experts)}} & \multicolumn{2}{c}{\textbf{arch6 (6 experts)}} \\
\cmidrule(lr){2-3} \cmidrule(lr){4-5}
\textbf{Method} & \multicolumn{1}{l}{\textbf{Macro AUROC}} & \multicolumn{1}{l}{\textbf{Params}$^{*}$} & \multicolumn{1}{l}{\textbf{Macro AUROC}} & \multicolumn{1}{l}{\textbf{Params}$^{*}$} \\
\midrule
\multicolumn{5}{l}{\textit{Core PS baselines}} \\
\psmethod{1}: Probability averaging & 0.868 & 0 & 0.682 & 0 \\
\psmethod{2}: Logistic regression & 0.879 & 9 & 0.786 & 13 \\
\psmethod{3}: MLP on logits & 0.901 & 706 & 0.786 & 962 \\
\psmethod{4}: XGBoost & 0.939 & - & 0.843 & - \\
\midrule
\multicolumn{5}{l}{\textit{Additional PS baselines}} \\
\psmethod{A}: Calibrated averaging & 0.860 & 4 & 0.748 & 6 \\
\psmethod{B}: Learned averaging weights & 0.836 & 4 & 0.787 & 6 \\
\psmethod{C}: CV-tuned logistic regression & 0.879 & 9 & 0.785 & 13 \\
\psmethod{D}: MLP on probabilities & 0.835 & 706 & 0.759 & 962 \\
\psmethod{E}: MLP on probabilities + early stopping & 0.875 & 706 & 0.763 & 962 \\
\psmethod{F}: MLP on logits + early stopping & 0.912 & 706 & 0.773 & 962 \\
\midrule
\multicolumn{5}{l}{\textit{Richer PS inputs}} \\
\psmethod{G}-LR: Disagreement + logistic regression & 0.892 & 15 & 0.790 & 28 \\
\psmethod{G}-XGB: Disagreement + XGBoost & \textbf{0.942} & - & 0.860 & - \\
\psmethod{H}-LR: Uncertainty summaries + logistic regression & 0.885 & 18 & 0.779 & 22 \\
\psmethod{H}-XGB: Uncertainty summaries + XGBoost & 0.935 & - & 0.844 & - \\
\psmethod{I}-LR: Quadratic interactions + logistic regression & 0.890 & 19 & 0.778 & 34 \\
\psmethod{I}-XGB: Quadratic interactions + XGBoost & 0.939 & - & \textbf{0.860} & - \\
\psmethod{J}-LR: Rank transformation + logistic regression & 0.869 & 5 & 0.771 & 7 \\
\psmethod{J}-XGB: Rank transformation + XGBoost & 0.932 & - & 0.832 & - \\
\psmethod{K}-LR: CV-selected input + logistic regression & 0.890 & 19 & 0.778 & 34 \\
\psmethod{K}-XGB: CV-selected input + XGBoost & 0.939 & - & 0.860 & - \\
\midrule
\multicolumn{5}{l}{\textit{FS comparators}} \\
\fsmethod{1}: Logistic regression on raw features & 0.965 & 4.1K & 0.966 & 10.0K \\
\fsmethod{2}: Learned projections + MLP & 0.980 & 329K & 0.969 & 3.5M \\
\midrule
\multicolumn{5}{l}{\textit{FS controls}} \\
\fsmethod{A}: Learned projections + linear classifier & 0.977 & 263K & - & - \\
\fsmethod{B}: PCA + logistic regression & 0.941 & 65 & - & - \\
\fsmethod{C}: Random projections + MLP & 0.969 & 264K & - & - \\
\fsmethod{D}: XGBoost on features & 0.959 & - & 0.954 & - \\

\bottomrule
\end{tabular}
}
\\[2pt]
{\scriptsize Standard split (overlap-affected). Bold selects the ceiling using unrounded AUROC; \psmethod{I}-XGB narrowly exceeds \psmethod{G}-XGB on arch6. \psmethod{K} selects map I on both pools. Stochastic results average predictions over eight seeds. $^{*}$Counts are per model, excluding frozen experts: fitted classifier/projection parameters, temperatures and averaging weights are included; standardization, CDF values, PCA and fixed random mappings are excluded. \fsmethod{B} therefore counts only its final classifier. Tree parameter counts are unreported; an AUROC dash means unevaluated. \fsmethod{2} uses $d=32/256$ for clip4/arch6; clip4 controls A/C use $d=64$ and B retains 64 total PCA components.}
\end{table}

\paragraph{Results and comparator.} The full-ladder ceilings are 0.860 for arch6 (\psmethod{I}-XGB) and 0.942 for clip4 (\psmethod{G}-XGB). \fsmethod{1} exceeds them by +0.105 $[+0.058,+0.158]$ and +0.024 $[+0.007,+0.043]$, respectively, winning all eight generators in each pool (sign-flip $p=0.008$). Standard-split selection uses one best method by macro test AUROC; LOGO selects separately per fold. Both are retrospective, with intervals conditional on the selected comparators (\S\ref{sec:psladder}).

\paragraph{Target-label diagnostic.} A separate diagnostic fits standardized, class-balanced logistic regression ($C=1$) on the experts' logits by five-fold CV within each generator's labeled test set, scoring each sample out of fold. Its macro AUROC is 0.952 on clip4 and 0.868 on arch6. Access to target labels and generator identity excludes it from the PS ceiling.

\paragraph{Raw-feature scaling check.} Fitting a \texttt{StandardScaler} on TRAIN and refitting \fsmethod{1} changes standard-split macro AUROC from 0.9657 to 0.9654 on arch6 and from 0.9654 to 0.9659 on clip4 (unrounded differences $-0.0003$, $+0.0004$). Individual-generator changes can be larger: clip4 HeyGen rises from 0.7773 to 0.8239. Thus macro stability here does not establish invariance across generators or under LOGO.

\begin{table}[!htbp]
\caption{Per-generator AUROC for selected methods on arch6 under the standard split (\fsmethod{2}: $d\!=\!256$).}
\label{tab:arch6}
\centering
{\footnotesize
\setlength{\tabcolsep}{5pt}
\begin{tabular}{lrrrrrrrr r}
\toprule
\textbf{Method} & \multicolumn{1}{l}{\textbf{WIR}} & \multicolumn{1}{l}{\textbf{CDiff}} & \multicolumn{1}{l}{\textbf{MJ6}} & \multicolumn{1}{l}{\textbf{HeyG}} & \multicolumn{1}{l}{\textbf{SG2}} & \multicolumn{1}{l}{\textbf{DFL}} & \multicolumn{1}{l}{\textbf{SG}} & \multicolumn{1}{l}{\textbf{SCLIP}} & \multicolumn{1}{l}{\textbf{Macro AUROC}} \\
\midrule
\psmethod{1}: Prob.\ avg. & 0.553 & 0.731 & 0.391 & 0.677 & 0.730 & 0.950 & 0.779 & 0.647 & 0.682 \\
\psmethod{4}: XGBoost & 0.609 & 0.884 & 0.866 & 0.698 & 0.925 & 0.938 & 0.923 & 0.900 & 0.843 \\
\fsmethod{1}: Raw-feat LogReg & 0.867 & 1.000 & 1.000 & 0.902 & 1.000 & 0.979 & 0.978 & 1.000 & 0.966 \\
\fsmethod{2}: MLP ($d\!=\!256$) & 0.848 & 1.000 & 1.000 & 0.933 & 1.000 & 0.988 & 0.986 & 1.000 & 0.969 \\
\bottomrule
\end{tabular}
}
\\[2pt]
{\scriptsize Standard split (overlap-affected); stochastic results use seed-averaged predictions. \psmethod{4} is the best of the original ten methods, not the full-ladder ceiling. WIR/ CDiff/ MJ6/ HeyG/ SG2/ DFL/ SG/ SCLIP abbreviate WhichIsReal/ CollabDiff/ MidJourney/ HeyGen/ StarGAN-v2/ DeepFaceLab/ StarGAN/ StyleCLIP. Rounding to 1.000 does not imply an exact AUROC of one.}
\end{table}

\FloatBarrier

\begin{table}[H]
\caption{Per-generator AUROC for selected methods on clip4 under the standard split (\fsmethod{2}: $d\!=\!32$).}
\label{tab:clip4}
\centering
{\footnotesize
\setlength{\tabcolsep}{5pt}
\begin{tabular}{lrrrrrrrr r}
\toprule
\textbf{Method} & \multicolumn{1}{l}{\textbf{WIR}} & \multicolumn{1}{l}{\textbf{CDiff}} & \multicolumn{1}{l}{\textbf{MJ6}} & \multicolumn{1}{l}{\textbf{HeyG}} & \multicolumn{1}{l}{\textbf{SG2}} & \multicolumn{1}{l}{\textbf{DFL}} & \multicolumn{1}{l}{\textbf{SG}} & \multicolumn{1}{l}{\textbf{SCLIP}} & \multicolumn{1}{l}{\textbf{Macro AUROC}} \\
\midrule
\psmethod{1}: Prob.\ avg. & 0.858 & 0.914 & 0.684 & 0.791 & 0.847 & 0.966 & 0.984 & 0.896 & 0.868 \\
\psmethod{4}: XGBoost & 0.885 & 0.998 & 0.969 & 0.698 & 0.993 & 0.978 & 0.991 & 0.997 & 0.939 \\
\fsmethod{1}: Raw-feat LogReg & 0.960 & 1.000 & 1.000 & 0.777 & 1.000 & 0.987 & 0.999 & 1.000 & 0.965 \\
\fsmethod{2}: MLP ($d\!=\!32$) & 0.953 & 1.000 & 1.000 & 0.902 & 1.000 & 0.986 & 0.999 & 1.000 & 0.980 \\
\bottomrule
\end{tabular}
}
\\[2pt]
{\scriptsize Standard split (overlap-affected). Comparator, abbreviations and rounding follow Table~\ref{tab:arch6}; the full macro ladder is in Table~\ref{tab:ladder}.}
\end{table}

\subsection*{C. \fsmethod{2} Robustness Checks}\label{app:C}

\paragraph{C.1. Projection dimension.} On clip4, we vary only the per-expert projection width $d\in\{32,64,256\}$, retaining Appendix A's settings. Macro AUROC varies by less than 0.001 despite increasing parameter counts. This is a standard-split sensitivity check, not evidence of equivalence across architectures or protocols.

\begin{table}[!htbp]
\caption{\fsmethod{2} projection-dimension sweep on clip4 under the standard split.}
\label{tab:dim_sweep}
\centering
\begin{tabular}{rrrrr}
\toprule
\multicolumn{1}{l}{\textbf{Per-expert $d$}} & \multicolumn{1}{l}{\textbf{Combined dimension}} & \multicolumn{1}{l}{\textbf{Macro AUROC}} & \multicolumn{1}{l}{\textbf{$\Delta$ vs PS ceiling}} & \multicolumn{1}{l}{\textbf{Params}} \\
\midrule
32 & 128 & 0.980 & +0.038 & 329K \\
64 & 256 & 0.981 & +0.039 & 526K \\
256 & 1,024 & 0.981 & +0.039 & 1.7M \\
\bottomrule
\end{tabular}
\\[2pt]
{\scriptsize Overlap-affected standard split. Macro AUROC uses seed-averaged predictions; parameters refer to one model. The comparator is \psmethod{G}-XGB (0.942); differences use unrounded values.}
\end{table}

\paragraph{C.2. Seeds and SELECT--TEST differences.} Table~\ref{tab:fs2_stability} reports individual-seed macro TEST AUROCs before prediction averaging and pooled SELECT AUROCs of the chosen checkpoints. TEST SD is at most 0.004 and the range at most 0.012; every seeded model exceeds its pool's standard-split PS ceiling. Prediction averaging adds approximately 0.002--0.006 relative to the mean per-seed macro AUROC.

\begin{table}[!htbp]
\caption{\fsmethod{2} per-seed performance and SELECT--TEST differences under the standard split.}
\label{tab:fs2_stability}
\centering
{\footnotesize
\setlength{\tabcolsep}{4pt}
\begin{tabular}{lrrrrrr}
\toprule
\textbf{Pool} & \multicolumn{1}{l}{\textbf{$d$}} & \multicolumn{1}{l}{\textbf{TEST mean}} & \multicolumn{1}{l}{\textbf{TEST SD}} & \multicolumn{1}{l}{\textbf{TEST min-max}} & \multicolumn{1}{l}{\textbf{SELECT mean}} & \multicolumn{1}{l}{\textbf{SELECT$-$TEST}} \\
\midrule
clip4 & 32  & 0.974 & 0.003 & 0.970-0.981 & 1.000 & $+0.026$ \\
clip4 & 64  & 0.975 & 0.004 & 0.969-0.981 & 1.000 & $+0.024$ \\
clip4 & 256 & 0.979 & 0.003 & 0.975-0.984 & 1.000 & $+0.021$ \\
arch6 & 256 & 0.966 & 0.003 & 0.963-0.975 & 0.999 & $+0.033$ \\
\bottomrule
\end{tabular}
}
\\[2pt]
{\scriptsize Overlap-affected standard split. TEST statistics summarize individual-seed macro AUROCs; SD uses divisor 8 and is not uncertainty for the prediction ensemble. SELECT averages pooled checkpoint AUROCs. Gaps use unrounded means; 1.000 can reflect rounding.}
\end{table}

SELECT--TEST differences of +0.021--+0.033 do not isolate overfitting: SELECT and TEST use different aggregation, SELECT chooses checkpoints, and content is repeated across partitions (\app{A}). These checks cover seed variation and projection width in the tested standard-split configurations; they neither exclude memorization nor establish an optimal feature combiner. LOGO supplies the primary transfer evidence.

\subsection*{D. Representational Bottleneck}\label{sec:bottleneck-appendix}\label{app:D}

\paragraph{D.1. Effective dimensionality.} We center concatenated features and logits without feature-wise standardization on the 17,925 pooled test rows, retaining their multiplicities. Randomized SVD (seed 42) retains 3,000 feature components and all but the smallest logit component. We report participation ratio
\[
\mathrm{PR}=\frac{(\sum_j\lambda_j)^2}{\sum_j\lambda_j^2}.
\]
Here, $\lambda_j$ are the retained covariance eigenvalues. We also report $k_{95}$ and $k_{99}$, the numbers of retained components accounting for 95\% and 99\% of their summed variance. These statistics depend on scaling, spectral truncation and sample composition; they do not measure task-relevant information directly.

\begin{table}[!htbp]
\caption{Representation dimensionality and retained-spectrum summaries, pooled across the eight test sets shared by standard-split and LOGO evaluations.}
\label{tab:bottleneck}
\centering
\begin{tabular}{llrrrr}
\toprule
\textbf{Pool} & \textbf{Representation} & \multicolumn{1}{l}{\textbf{Input dimension}} & \multicolumn{1}{l}{\textbf{PR}} & \multicolumn{1}{l}{\textbf{$k_{95}$}} & \multicolumn{1}{l}{\textbf{$k_{99}$}} \\
\midrule
arch6 & Concatenated logits & 12 & 2.55 & 5 & 6 \\
arch6 & Concatenated features & 9,984 & 47.5 & 877 & 1,458 \\
clip4 & Concatenated logits & 8 & 1.87 & 3 & 5 \\
clip4 & Concatenated features & 4,096 & 44.8 & 1,324 & 2,377 \\
\bottomrule
\end{tabular}
\\[2pt]
{\scriptsize Statistics use the retained spectrum, without feature-wise standardization, and retain repeated rows.}
\end{table}

Dividing feature PR by logit PR gives approximately $18.6$ on arch6 ($47.5/2.55$) and $24.0$ on clip4 ($44.8/1.87$). The latter has the larger PR gap but smaller standard-split feature advantage, so dimensionality alone does not explain the performance pattern.

\begin{figure}[htbp]
\centering
\includegraphics[width=\textwidth]{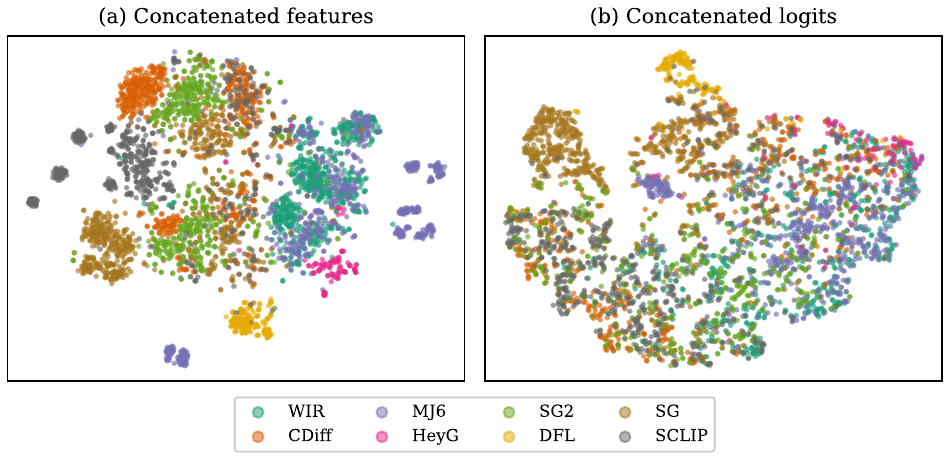}
\caption{t-SNE views of clip4 representations: (a) concatenated features and (b) concatenated logits, colored by test generator. The visualization uses 4,000 cached rows from the test sets shared by standard-split and LOGO evaluations; generator abbreviations follow Table~\ref{tab:arch6}. The panels illustrate local structure qualitatively. Their axes and inter-cluster distances are not directly comparable.}
\label{fig:tsne}
\end{figure}

\paragraph{D.2. t-SNE visualization.} Figure~\ref{fig:tsne} shows clip4's 4,096 feature values and eight logits using 4,000 cached rows (seed 42; up to 625 per generator). Both embeddings use perplexity 30, 1,000 iterations, seed 42, no feature-wise standardization and remaining scikit-learn defaults~\cite{vandermaaten2008}. Generator-dominated regions in feature space are qualitative structure, not evidence of real-versus-fake information or a quantitative comparison of the interfaces.

\begin{figure}[htbp]
\centering
\includegraphics[width=0.8\textwidth]{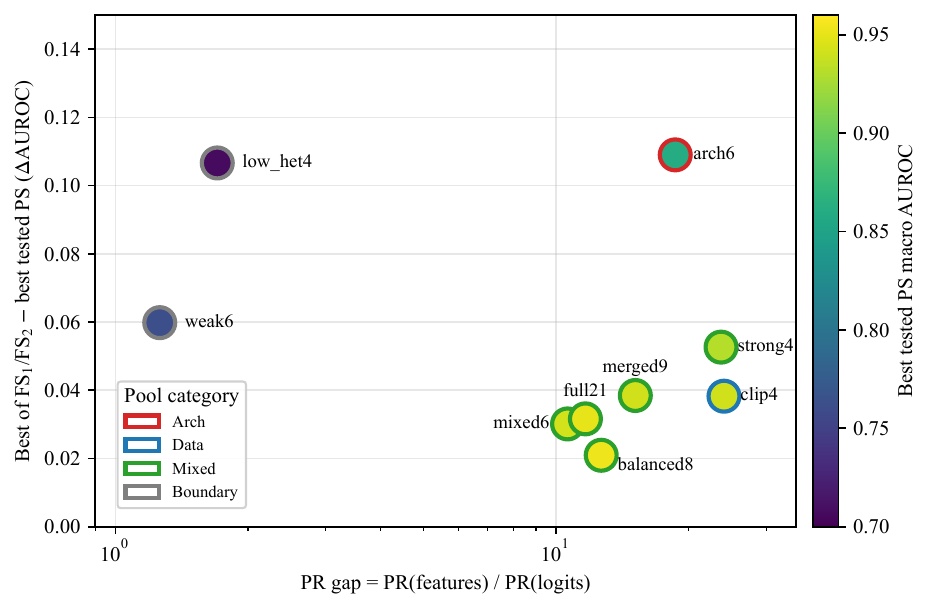}
\caption{Exploratory relationship between the PR gap (horizontal axis, logarithmic scale) and $\max(\Delta_1, \Delta_2)$ on the overlap-affected standard split. The vertical value takes the better of \fsmethod{1} and \fsmethod{2} in each pool, chosen retrospectively. Fill color indicates the best tested PS macro AUROC; outline color indicates the pool category: one heterogeneity axis varied (Arch, Data), both axes varied (Mixed), or a boundary pool. The full PS ladder is evaluated only on arch6 and clip4; the other seven pools use narrower sets of PS methods. The figure does not isolate a causal relationship between representation dimensionality and performance.}
\label{fig:bottleneck_scatter}
\end{figure}

\paragraph{D.3. Cross-pool scatter.} Figure~\ref{fig:bottleneck_scatter} plots the PR ratio against $\max(\Delta_1,\Delta_2)$ from Table~\ref{tab:fs1_pools}, retrospectively choosing the better FS combiner. Pools differ in expert composition and PS-ladder coverage; the scatter therefore does not isolate an effect of representation dimensionality.

\paragraph{D.4. Width-matched controls.} For arch6, two values per expert match the 12-value prediction interface. Label-fitted values come from a logistic direction separating real from fake, followed by a second logistic direction fitted after projecting out the first. PCA instead retains two highest-variance directions per expert. Both mappings and the final \fsmethod{1} combiner are fitted on TRAIN: once on the standard split and separately within each LOGO fold.

\begin{table}[ht]
\caption{Combiner comparisons and arch6 input-width controls: macro AUROC over the eight held-out generators.}
\label{tab:rq2}
\centering
{\footnotesize
\setlength{\tabcolsep}{4pt}
\begin{tabular}{lrrrrrr}
\toprule
 & & & \multicolumn{2}{l}{\textbf{Standard split}} & \multicolumn{2}{l}{\textbf{LOGO}} \\
\cmidrule(lr){4-5} \cmidrule(lr){6-7}
\textbf{Combiner} & \multicolumn{1}{l}{\textbf{Inputs (arch6)}} & \multicolumn{1}{l}{\textbf{Params (arch6)}} & \multicolumn{1}{l}{\textbf{arch6}} & \multicolumn{1}{l}{\textbf{clip4}} & \multicolumn{1}{l}{\textbf{arch6}} & \multicolumn{1}{l}{\textbf{clip4}} \\
\midrule
\psmethod{4}: XGBoost on logits & 12 & - & 0.843 & 0.939 & 0.727 & 0.730 \\
\fsmethod{1}: Logistic regression on raw features & 9,984 & 10.0K & 0.966 & 0.965 & 0.898 & 0.846 \\
\fsmethod{2}: Learned projections + MLP & 9,984 & 3.5M & 0.969 & 0.980 & 0.896 & 0.893 \\
\fsmethod{D}: XGBoost on features & 9,984 & - & 0.954 & 0.959 & 0.869 & 0.852 \\
\fsmethod{1} on 12 PCA values & 12 & 13$^{*}$ & 0.740 & - & 0.730 & - \\
\fsmethod{1} on 12 label-fitted values & 12 & 13$^{*}$ & 0.958 & - & 0.880 & - \\
\bottomrule
\end{tabular}
}
\\[2pt]
{\scriptsize Standard split is overlap-affected. \psmethod{4} is the control comparator, not the full-ladder ceiling. Counts exclude experts and refer to one model; \fsmethod{2} uses seed-averaged predictions. $^{*}$The 13 parameters count only the final classifier, excluding learned mappings. Dashes denote unreported entries.}
\end{table}

On the standard split, label-fitted values preserve most of \fsmethod{1}'s advantage over \psmethod{4} (0.958 versus 0.966 and 0.843); the two-value PCA control reaches 0.740 (versus \psmethod{4}, sign-flip $p=0.109$). Under LOGO, the label-fitted control exceeds \psmethod{4} on all eight generators but trails \fsmethod{1} by $-0.018$ $[-0.043,-0.001]$. The PCA difference from \psmethod{4} is +0.003 $[-0.026,+0.033]$.

PCA widths of one, four and eight values per expert give standard-split AUROCs 0.783, 0.766 and 0.818, and LOGO AUROCs 0.772, 0.795 and 0.828. The two-value control is the lowest under both protocols: the trend is not monotone. Removing training duplicates under LOGO changes the two-value PCA difference to $-0.032$ $[-0.062,+0.001]$ and the label-fitted difference from \fsmethod{1} to $-0.023$ $[-0.056,+0.002]$ (\app{A}); neither interval establishes equivalence.

On clip4, eight label-fitted values reach 0.971 versus \fsmethod{1}'s 0.965 on the standard split, and 0.872 versus 0.846 under LOGO (difference +0.026 $[-0.004,+0.076]$). PCA reaches 0.844 and 0.831, against \psmethod{4}'s 0.939 and 0.730. These controls require full-feature access and fitted mappings beyond the final classifier. Matching input width does not match total learned capacity or establish that dimensionality is irrelevant.

\subsection*{E. Leave-One-Expert-Out Robustness}\label{app:E}

We remove each expert and retrain \fsmethod{2}, retaining the standard split, eight seeds, optimization and SELECT checkpoint rule (\app{A}), with $d=32$ for clip4 and $d=256$ for arch6. AUROCs use seed-averaged predictions. The removal effect is full-pool minus retrained reduced-pool AUROC; it measures adaptation through retraining, not failure at inference without retraining.

\begin{table}[ht]
\caption{\fsmethod{2} performance after removing one expert and retraining, under the standard split.}
\label{tab:loo}
\centering
\begin{tabular}{llrr}
\toprule
\textbf{Pool} & \textbf{Expert removed} & \multicolumn{1}{l}{\textbf{Macro AUROC after retraining}} & \multicolumn{1}{l}{\textbf{Full $-$ reduced AUROC}} \\
\midrule
clip4 & None: full pool & 0.980 & - \\
clip4 & fs-clip-large & 0.970 & +0.010 \\
clip4 & fr-clip-large & 0.976 & +0.004 \\
clip4 & efs-clip-large & 0.970 & +0.010 \\
clip4 & df40-clip-large & 0.977 & +0.003 \\
\midrule
arch6 & None: full pool & 0.969 & - \\
arch6 & fs-clip-base & 0.969 & +0.001 \\
arch6 & fs-clip-large & 0.943 & +0.026 \\
arch6 & fs-xception & 0.966 & +0.003 \\
arch6 & fs-srm & 0.968 & +0.001 \\
arch6 & fs-recce & 0.968 & +0.001 \\
arch6 & fs-rfm & 0.969 & 0.000 \\
\bottomrule
\end{tabular}
\\[2pt]
{\scriptsize Overlap-affected standard split; differences use unrounded, seed-averaged AUROCs. Removing fs-rfm improves AUROC by about 0.000067, rounding to a removal effect of 0.000. Dashes mark full pools.}
\end{table}

On clip4, removal reduces macro AUROC by approximately 0.003--0.010. On arch6, removing fs-clip-large causes the largest decrease (0.026); every other removal changes AUROC by at most 0.003. These conditional effects change both available representations and combiner size and have no uncertainty intervals. They do not establish expert redundancy or additive contributions. PS baselines were not refitted, so this overlap-affected \fsmethod{2} diagnostic does not establish robustness of the FS--PS advantage.

\subsection*{F. Leave-One-Generator-Out Control (LOGO)}\label{app:F}

\paragraph{LOGO protocol.} Each fold trains combiners on seven generator-specific test sets and evaluates on the eighth. Before the stratified 85/15 TRAIN/SELECT split (seed 1024), it excludes rows whose features match held-out test content. Seven rows are removed in each of CollabDiff and StyleCLIP, corresponding to shared real images. Folds retain about 11.8K--15K TRAIN and 2K--2.7K SELECT rows. This controls identified combiner train/test reuse, but not repeated content within training/selection, re-encoded copies or frozen-expert exposure (\app{A}).

\begin{table}[ht]
\caption{Fold-wise LOGO results on arch6.}
\label{tab:logo}
\centering
\begin{tabular}{lrlrrrr}
\toprule
\textbf{Held-out generator} & \multicolumn{1}{l}{\textbf{PS ceiling}} & \textbf{Selected PS method} & \multicolumn{1}{l}{\textbf{\fsmethod{1}}} & \multicolumn{1}{l}{\textbf{$\Delta_1$}} & \multicolumn{1}{l}{\textbf{\fsmethod{2}}} & \multicolumn{1}{l}{\textbf{$\Delta_2$}} \\
\midrule
WhichIsReal & 0.587 & \psmethod{B} & 0.814 & +0.227 & 0.832 & +0.245 \\
CollabDiff & 0.884 & \psmethod{C} & 0.999 & +0.114 & 0.999 & +0.114 \\
MidJourney & 0.713 & \psmethod{C} & 0.868 & +0.155 & 0.832 & +0.119 \\
HeyGen & 0.769 & \psmethod{H}-LR & 0.771 & +0.001 & 0.731 & $-$0.039 \\
StarGAN-v2 & 0.815 & \psmethod{B} & 0.853 & +0.038 & 0.862 & +0.048 \\
DeepFaceLab & 0.960 & \psmethod{B} & 0.916 & $-$0.045 & 0.937 & $-$0.023 \\
StarGAN & 0.951 & \psmethod{B} & 0.970 & +0.020 & 0.979 & +0.028 \\
StyleCLIP & 0.859 & \psmethod{C} & 0.993 & +0.134 & 0.997 & +0.137 \\
\midrule
Macro AUROC & 0.817 & - & 0.898 & +0.081 & 0.896 & +0.079 \\
\bottomrule
\end{tabular}
\\[2pt]
{\scriptsize The full-ladder PS ceiling is selected on TEST separately per fold. \psmethod{B} fits averaging weights; \psmethod{C} is CV-tuned logistic regression; \psmethod{H}-LR uses uncertainty/MI inputs. $\Delta_i=\mathrm{FS}_i-\mathrm{PS\ ceiling}$ uses unrounded scores; \fsmethod{2} uses seed-averaged predictions.}
\end{table}

\fsmethod{1} exceeds the ceiling in seven folds; \fsmethod{2} does so in six, losing on HeyGen and DeepFaceLab. The lower macro AUROC than on the standard split combines changes in training size, composition and overlap, whose effects are not separated here.

\paragraph{Sensitivity to small test sets.} HeyGen and DeepFaceLab have 101 and 149 rows. This exploratory summary excludes them, equally averaging existing differences over the six remaining generators with $n\geq500$; no models are refitted. \fsmethod{1} wins all six on arch6. All eight generators remain primary: exclusion changes the evaluated tasks and does not establish why results differ.

\begin{center}
{\footnotesize
\begin{tabular}{lrr}
\toprule
\textbf{Pool} & \multicolumn{1}{l}{\textbf{$\Delta$, all eight}} & \multicolumn{1}{l}{\textbf{$\Delta$, six with $n\!\geq\!500$}} \\
\midrule
arch6 & +0.081 & +0.115 \\
clip4 & $-$0.074 & $-$0.010 \\
strong4 & $-$0.060 & +0.001 \\
weak6 & $-$0.102 & $-$0.059 \\
\bottomrule
\end{tabular}
}
\\[2pt]
{\scriptsize $\Delta$ = \fsmethod{1} $-$ PS ceiling under LOGO; differences are calculated before rounding.}
\end{center}

\paragraph{Inferential views.} Table~\ref{tab:logo_inferential} reports two-sided sign and Wilcoxon signed-rank tests, a sign-flip test over all $2^8$ patterns, and a percentile interval from 10,000 paired fold resamples~\cite{efron1979}. These analyses treat folds as independent, although two LOGO folds share six training generators; this can underestimate uncertainty~\cite{bengio2004}. Sign-flip enumeration is exact under its null assumptions, including independent, symmetric differences, not a remedy for dependence. All summaries condition on the retrospectively selected PS comparator.

\begin{table}[ht]
\caption{Inferential summaries for arch6 LOGO. Tests are two-sided; $\Delta$ is the mean fold difference.}
\label{tab:logo_inferential}
\centering
{\footnotesize
\begin{tabular}{lrrrrr}
\toprule
\textbf{Comparison} & \multicolumn{1}{l}{\textbf{$\Delta$}} & \multicolumn{1}{l}{\textbf{Sign-test $p$}} & \multicolumn{1}{l}{\textbf{Wilcoxon $p$}} & \multicolumn{1}{l}{\textbf{Sign-flip $p$}} & \multicolumn{1}{l}{\textbf{Bootstrap 95\% CI}} \\
\midrule
\fsmethod{1} versus PS ceiling & +0.081 & 0.070 & 0.055 & 0.055 & $[+0.024, +0.140]$ \\
\fsmethod{2} versus PS ceiling & +0.079 & 0.289 & 0.055 & 0.055 & $[+0.022, +0.141]$ \\
\bottomrule
\end{tabular}
}
\\[2pt]
{\scriptsize Nominal percentile-bootstrap intervals; comparator-selection, dependence and small-sample qualifications apply as discussed in the text.}
\end{table}

Both intervals exclude zero, but none of the three tests reaches $p<0.05$. \fsmethod{1} remains the primary inferential target; the positive estimated advantage has limited inferential precision.

\paragraph{Small-sample diagnostic.} An illustrative simulation uses 10,000 sets of eight independent Normal differences, with 100,000-set precision checks near bootstrap detection thresholds. At SD 0.08, the nominal 95\% percentile interval excludes zero in approximately 11\% of null simulations; the sign-flip test stays near 5\% and requires approximately +0.095 for 80\% power. The two-sided sign test needs eight same-direction differences to reach $p<0.05$. This simulation does not model LOGO dependence or estimate the study's actual error rate or power. Read bootstrap intervals alongside the paired tests. Full detection-threshold outputs remain in the release's statistical analysis.

\paragraph{Weighting robustness.} Table~\ref{tab:weighting} compares the equal-generator mean with $\sum_g n_g\mathrm{AUROC}_g/\sum_g n_g$, retaining each selected PS comparator. The latter weights within-generator AUROCs by cached-row counts, including repeated images; it is not pooled-prediction AUROC. All six reported differences remain positive, although their magnitudes change. Equal-generator macro AUROC remains primary.

\begin{table}[ht]
\caption{Sensitivity to generator weighting. $\Delta$ = FS $-$ selected PS comparator.}
\label{tab:weighting}
\centering
{\footnotesize
\begin{tabular}{llrrrrrr}
\toprule
 & & \multicolumn{3}{l}{\textbf{Unweighted macro}} & \multicolumn{3}{l}{\textbf{Sample-weighted macro}} \\
\cmidrule(lr){3-5} \cmidrule(lr){6-8}
\textbf{Pool and protocol} & \textbf{Variant} & \multicolumn{1}{l}{\textbf{FS}} & \multicolumn{1}{l}{\textbf{PS}} & \multicolumn{1}{l}{\textbf{$\Delta$}} & \multicolumn{1}{l}{\textbf{FS}} & \multicolumn{1}{l}{\textbf{PS}} & \multicolumn{1}{l}{\textbf{$\Delta$}} \\
\midrule
arch6, standard & \fsmethod{1} & 0.966 & 0.860 & +0.105 & 0.980 & 0.896 & +0.083 \\
arch6, standard & \fsmethod{2} & 0.969 & 0.860 & +0.109 & 0.979 & 0.896 & +0.083 \\
clip4, standard & \fsmethod{1} & 0.965 & 0.942 & +0.024 & 0.994 & 0.979 & +0.015 \\
clip4, standard & \fsmethod{2} & 0.980 & 0.942 & +0.038 & 0.994 & 0.979 & +0.015 \\
arch6, LOGO & \fsmethod{1} & 0.898 & 0.817 & +0.081 & 0.922 & 0.827 & +0.095 \\
arch6, LOGO & \fsmethod{2} & 0.896 & 0.817 & +0.079 & 0.925 & 0.827 & +0.098 \\
\bottomrule
\end{tabular}
}
\\[2pt]
{\scriptsize Standard-split results are overlap-affected and use one macro-AUROC-selected PS method; LOGO selects per fold. Differences use unrounded scores.}
\end{table}

\paragraph{Richer prediction-space inputs inside LOGO.} Each fold refits the four maps of \app{A}. Their logistic heads standardize using TRAIN and use $C=1$ without class weights. XGBoost selects by three-fold TRAIN CV: the initial grid has 100/200 trees, depths 2/3/4, learning rate 0.1 and row/column subsampling 0.8; the expanded grid adds 500/1,000 trees, depths 6/8/10 and learning rate 0.05. Each fold's ceiling considers 22 candidates: ten original methods plus logistic regression and the selections from both XGBoost grids for each map. \psmethod{K} is not separately fitted because its maps are represented. Appendix A reports the standard-head sensitivity check.

On arch6, the richer-input \psmethod{H}-LR sets the HeyGen ceiling. Expanding the ten-method ladder changes the \fsmethod{1} margin from +0.082 $[+0.026,+0.140]$ to +0.081 $[+0.024,+0.140]$ and sign-flip $p$ from 0.047 to 0.055. Other pools' mean margins move by less than 0.003. Thus the point estimate is stable, while crossing the 0.05 threshold depends on baseline coverage.

\paragraph{Comparator-selection and ladder sensitivity.} Selecting one PS method by its mean AUROC across folds, rather than separately per fold, gives \fsmethod{1} margins +0.111, $-0.044$, $-0.038$ and $-0.028$ for arch6, clip4, strong4 and weak6. Both rules select retrospectively on TEST and neither is a deployment rule. The directions agree, but weak6's magnitude is particularly sensitive. \fsmethod{1} also trails probability averaging on clip4 (0.846 versus 0.868) and strong4 (0.850 versus 0.869).

No tree candidate sets a full-ladder fold ceiling in these four pools. Restricting PS to \psmethod{1}--\psmethod{3} inflates the feature margin by at most +0.026. The Minimum tier of Table~\ref{tab:tiers}, the better of \psmethod{1} and \psmethod{4}, gives +0.154 rather than +0.081 on arch6, illustrating the value of broader baseline coverage.

Table~\ref{tab:audit}'s +0.025 standard--LOGO contrast also changes with selection: using the same rule in both protocols gives $-0.006$ for one best method (+0.105 versus +0.111) or +0.011 for per-generator selection (+0.091 versus +0.081). These contrasts combine comparator, training-data and overlap differences; none isolates reuse causally.

\paragraph{Strict path-disjoint refit.} Removing every standard TRAIN/SELECT row whose leading path component names a test generator removes exactly the same-path test matches: 9,521 TRAIN and 2,365 SELECT rows. Real TRAIN rows fall from 5,438 to 586 and source families from eight to one, leaving FaceForensics++ originals~\cite{rossler2019faceforensics}. Refitting gives \fsmethod{1} 0.536 versus PS 0.764, a margin of $-0.228$.

A matched control removes the same counts within each class and partition at random; its mean margin across three seeds is +0.149. The strict-minus-random contrast is $-0.377$, but source coverage remains unmatched and random removal retains overlap. Even StarGAN, with no identified path reuse, falls from 0.978 to 0.508 under strict removal (random removal: 0.963--0.974). This is a coverage-confounded stress test, not an estimate of overlap's effect; LOGO remains primary.

\begin{table}[ht]
\caption{arch6 results under four evaluation conditions.}
\label{tab:strict}
\centering
{\footnotesize
\setlength{\tabcolsep}{4pt}
\begin{tabular}{lrl>{\raggedright\arraybackslash}p{5cm}}
\toprule
\textbf{Condition} & \multicolumn{1}{l}{\textbf{\fsmethod{1} $-$ PS comparator}} & \textbf{Bootstrap 95\% CI} & \textbf{Interpretation} \\
\midrule
Standard split & +0.105 & $[+0.058, +0.158]$ & Overlap-affected; secondary \\
LOGO & +0.081 & $[+0.024, +0.140]$ & Primary; seven of eight folds won \\
Strict path-disjoint refit & $-$0.228 & $[-0.322, -0.130]$ & \fsmethod{1} 0.536, PS 0.764; substantial coverage loss \\
Matched random removal & +0.149 & - & Mean across three removal seeds; sample and class counts matched \\
\bottomrule
\end{tabular}
}
\\[2pt]
{\scriptsize Strict/random removal modifies standard-split TRAIN/SELECT while retaining its test sets. Standard/LOGO use full PS ladders; strict/random use the original ten methods (standard margin +0.123 $[+0.078,+0.174]$ on that ladder). Selection is once by macro AUROC except per-fold LOGO. Intervals are comparator-conditional paired generator/fold bootstraps. Random removal is a three-seed mean without a combined interval; the arms do not isolate overlap.}
\end{table}

\subsection*{G. Boundary Conditions Across Expert Pools}\label{app:G}

Table~\ref{tab:fs1_pools} reports nine pools on the overlap-affected standard split. Beyond arch6 and clip4, weak6 and low\_het4 are boundary pools; the other five vary both architecture and training regime. The full PS ladder is used only for arch6 and clip4. Thus the \fsmethod{1} margin range +0.011 to +0.107 combines pool composition and baseline coverage and does not isolate diversity. Under LOGO, the four evaluated pools instead range from $-0.102$ to +0.081 (Table~\ref{tab:main}).

\begin{table}[ht]
\caption{Standard-split results across nine expert pools.}
\label{tab:fs1_pools}
\centering
{\footnotesize
\setlength{\tabcolsep}{4pt}
\begin{tabular}{lrlrrrrr}
\toprule
\textbf{Pool} & \multicolumn{1}{l}{\textbf{N}} & \textbf{Composition} & \multicolumn{1}{l}{\textbf{PS comparator}} & \multicolumn{1}{l}{\textbf{\fsmethod{1}}} & \multicolumn{1}{l}{\textbf{$\Delta_1$}} & \multicolumn{1}{l}{\textbf{\fsmethod{2}}} & \multicolumn{1}{l}{\textbf{$\Delta_2$}} \\
\midrule
clip4 & 4 & CLIP-large, four training regimes & 0.942 & 0.965 & +0.024 & 0.980 & +0.038 \\
arch6 & 6 & Six detector variants, face-swap-trained & 0.860 & 0.966 & +0.105 & 0.969 & +0.109 \\
mixed6 & 6 & Mixed architectures and training regimes & 0.942 & 0.967 & +0.025 & 0.972 & +0.030 \\
merged9 & 9 & clip4 $\cup$ arch6; shared expert counted once & 0.941 & 0.972 & +0.030 & 0.980 & +0.038 \\
strong4 & 4 & Four high-performing CLIP-based experts & 0.931 & 0.973 & +0.042 & 0.984 & +0.053 \\
balanced8 & 8 & Four regimes $\times$ two architectures & 0.953 & 0.964 & +0.011 & 0.974 & +0.021 \\
full21 & 21 & All available experts in the study & 0.951 & 0.972 & +0.021 & 0.982 & +0.032 \\
weak6 & 6 & Six lowest-performing individual experts & 0.762 & 0.822 & +0.060 & 0.784 & +0.022 \\
low\_het4 & 4 & Face-swap-trained Xception-family variants & 0.708 & 0.814 & +0.107 & 0.780 & +0.073 \\
\bottomrule
\end{tabular}
}
\\[2pt]
{\scriptsize Overlap-affected standard split; \fsmethod{2} uses seed-averaged predictions. PS is the full ladder on arch6/clip4 and \psmethod{4} elsewhere. Other pools' margins are upper bounds on their full-ladder margins: stronger PS candidates could reduce them. Differences use unrounded scores.}
\end{table}

\paragraph{Pool definitions.} weak6 comprises fr-recce, fr-xception, fr-srm, fr-rfm, fs-recce and fs-xception, the six lowest-scoring experts on these test generators. low\_het4 contains face-swap-trained Xception, SRM, RECCE and RFM. mixed6 combines df40-clip-large, fs-xception, fr-srm, efs-recce, efs-clip-base and fr-rfm; merged9 is the union of the primary pools. strong4 contains df40-clip-large, efs-clip-large, fs-clip-large and efs-clip-base. It is not strictly the top four: df40-clip-base scores 0.788 versus fs-clip-large's 0.787. balanced8 crosses CLIP-large/Xception with four training regimes; full21 uses all available experts, excluding unavailable df40-srm, df40-recce and df40-rfm.

weak6 and strong4 are retrospective selections on the same generators used for evaluation. Aggregate-trained experts have the exposure limitations of \app{A}. Pool comparisons also vary identity, size and individual performance, so the table provides descriptive boundary cases rather than a controlled effect of heterogeneity.

\paragraph{Baseline-strength example.} On weak6, the best of \psmethod{1}--\psmethod{3} reaches 0.590, while \psmethod{4} reaches 0.762. \fsmethod{1}'s apparent advantage is therefore +0.232 versus +0.060: the +0.172 inflation in Table~\ref{tab:audit}. Positive standard-split margins do not establish transfer: under LOGO, \fsmethod{1} trails the ceilings on weak6 and strong4. Further pool diagnostics remain in the release results.

\subsection*{H. Individual Expert Performance and AUPRC}\label{app:H}

\paragraph{H.1. Individual experts.} Within arch6, fs-clip-large has the highest macro AUROC (0.787). \fsmethod{1} reaches 0.966 on the standard split and 0.898 under LOGO, differences +0.179 and +0.111. These contextual comparisons do not replace trained PS baselines, since \fsmethod{1} also receives supervised combiner training.

\begin{table}[ht]
\caption{Individual expert AUROC on the eight test sets shared by standard-split and LOGO evaluations.}
\label{tab:experts}
\centering
{\footnotesize
\begin{tabular}{lrrrrrrrr r}
\toprule
\textbf{Expert} & \multicolumn{1}{l}{\textbf{WIR}} & \multicolumn{1}{l}{\textbf{CDiff}} & \multicolumn{1}{l}{\textbf{MJ6}} & \multicolumn{1}{l}{\textbf{HeyG}} & \multicolumn{1}{l}{\textbf{SG2}} & \multicolumn{1}{l}{\textbf{DFL}} & \multicolumn{1}{l}{\textbf{SG}} & \multicolumn{1}{l}{\textbf{SCLIP}} & \multicolumn{1}{l}{\textbf{Macro AUROC}} \\
\midrule
fs-clip-base & 0.435 & 0.686 & 0.504 & 0.576 & 0.745 & 0.927 & 0.895 & 0.743 & 0.689 \\
fs-clip-large & 0.591 & 0.809 & 0.706 & 0.682 & 0.815 & 0.960 & 0.950 & 0.784 & 0.787 \\
fs-xception & 0.535 & 0.547 & 0.385 & 0.384 & 0.616 & 0.865 & 0.577 & 0.426 & 0.542 \\
fs-srm & 0.538 & 0.698 & 0.299 & 0.507 & 0.616 & 0.854 & 0.606 & 0.572 & 0.586 \\
fs-recce & 0.509 & 0.491 & 0.297 & 0.556 & 0.598 & 0.883 & 0.580 & 0.400 & 0.539 \\
fs-rfm & 0.496 & 0.798 & 0.287 & 0.721 & 0.570 & 0.905 & 0.652 & 0.705 & 0.642 \\
fr-clip-base & 0.449 & 0.625 & 0.526 & 0.613 & 0.436 & 0.879 & 0.755 & 0.281 & 0.570 \\
fr-clip-large & 0.479 & 0.539 & 0.918 & 0.738 & 0.643 & 0.896 & 0.891 & 0.460 & 0.695 \\
fr-xception & 0.323 & 0.050 & 0.462 & 0.456 & 0.456 & 0.714 & 0.492 & 0.006 & 0.370 \\
fr-srm & 0.357 & 0.081 & 0.512 & 0.582 & 0.435 & 0.792 & 0.473 & 0.044 & 0.409 \\
fr-recce & 0.278 & 0.064 & 0.316 & 0.585 & 0.374 & 0.736 & 0.530 & 0.005 & 0.361 \\
fr-rfm & 0.325 & 0.033 & 0.514 & 0.597 & 0.423 & 0.761 & 0.407 & 0.009 & 0.384 \\
efs-clip-base & 0.825 & 0.982 & 0.539 & 0.562 & 0.821 & 0.747 & 0.945 & 0.940 & 0.795 \\
efs-clip-large & 0.925 & 0.999 & 0.626 & 0.609 & 0.946 & 0.739 & 0.994 & 0.997 & 0.854 \\
efs-xception & 0.772 & 0.997 & 0.470 & 0.327 & 0.676 & 0.491 & 0.776 & 0.980 & 0.686 \\
efs-srm & 0.794 & 0.997 & 0.337 & 0.358 & 0.702 & 0.554 & 0.769 & 0.983 & 0.687 \\
efs-recce & 0.757 & 0.979 & 0.439 & 0.515 & 0.724 & 0.570 & 0.768 & 0.965 & 0.715 \\
efs-rfm & 0.622 & 0.979 & 0.552 & 0.313 & 0.636 & 0.595 & 0.730 & 0.966 & 0.674 \\
df40-xception & 0.632 & 0.675 & 0.269 & 0.723 & 0.569 & 0.835 & 0.722 & 0.495 & 0.615 \\
df40-clip-base & 0.689 & 0.942 & 0.545 & 0.689 & 0.781 & 0.925 & 0.912 & 0.822 & 0.788 \\
df40-clip-large & 0.869 & 0.983 & 0.697 & 0.747 & 0.911 & 0.947 & 0.971 & 0.949 & 0.884 \\
\bottomrule
\end{tabular}
}
\\[2pt]
{\scriptsize Scores use fake-minus-real logits without direction reversal; macro AUROC equally weights generators. Prefixes fs/fr/efs/df40 mean face swap/reenactment/entire-face synthesis/DF40 aggregate. Generator abbreviations follow Table~\ref{tab:arch6}; expert exposure is described in \app{A}.}
\end{table}

\paragraph{H.2. Average precision.} We use scikit-learn's step-wise average precision, denoted AUPRC, with fake as positive, averaging equally across generators at their observed prevalences. Under arch6 LOGO, \fsmethod{1}'s macro AUPRC advantage is +0.080 versus each fold's AUROC-selected PS method, and +0.065 after training-duplicate removal (\app{A}). Comparators are not reselected by AUPRC.

On the overlap-affected standard split, \fsmethod{1} exceeds \psmethod{4} by +0.103 macro AUPRC on arch6 and +0.033 on clip4 (AUROC differences +0.123 and +0.026). These secondary checks use \psmethod{4}, not the full-ladder standard-split ceilings. Class prevalence and small test sets qualify their interpretation.

\begin{table}[ht]
\caption{Class prevalence in the cached test sets shared by standard-split and LOGO evaluations.}
\label{tab:prevalence}
\centering
{\footnotesize
\begin{tabular}{lrrr}
\toprule
\textbf{Generator} & \multicolumn{1}{l}{\textbf{Test rows}} & \multicolumn{1}{l}{\textbf{Fake-labeled rows}} & \multicolumn{1}{l}{\textbf{Fake-class fraction}} \\
\midrule
WhichIsReal & 2,001 & 1,001 & 0.500 \\
CollabDiff & 2,000 & 1,000 & 0.500 \\
MidJourney & 1,882 & 877 & 0.466 \\
HeyGen & 101 & 51 & 0.505 \\
StarGAN-v2 & 4,000 & 1,998 & 0.500 \\
DeepFaceLab & 149 & 100 & 0.671 \\
StarGAN & 3,968 & 1,984 & 0.500 \\
StyleCLIP & 3,824 & 2,824 & 0.738 \\
\bottomrule
\end{tabular}
}
\\[2pt]
{\scriptsize Counts are cached rows/labels, including repeated images and the integrity issues in \app{A}, not distinct images. Fractions use unrounded counts.}
\end{table}

\subsection*{I. Further Checks: Output Sufficiency, Speech, and Cost}\label{app:I}

\paragraph{I.1. Three-class tasks.} These exploratory comparisons predict three forgery types from either newly fitted three-class heads or the experts' original binary outputs. Outer folds hold out generator families from all classes, grouping families with detected shared content. Combiners fit TRAIN and use SELECT for method/checkpoint selection.

\textit{Task-specific heads.} On frozen, binary-trained arch6 backbones, we fit class-balanced multinomial logistic heads ($C=1$, L-BFGS, maximum 500 iterations) on standardized TRAIN features for GAN synthesis, diffusion and reenactment. Five-fold cross-fitting supplies PS training predictions, grouping repeated rows but not holding out generator families internally. Final heads refit all outer TRAIN to supply SELECT/TEST outputs. Macro one-versus-rest (OvR) AUROC averages equally over classes, then over three outer folds.

PS4-K3 tunes trees in a separate row-stratified CV on the cross-fitted TRAIN logits. Repeated-content groups cross these tuning partitions, and training predictions can come from heads fitted on tuning-validation rows. These dependencies are internal to outer TRAIN.

\begin{table}[ht]
\caption{Three-class heads on frozen arch6 backbones, evaluated over three outer folds holding out generator families.}
\label{tab:k3_heads}
\centering
{\footnotesize
\begin{tabular}{lr}
\toprule
\textbf{Method} & \multicolumn{1}{l}{\textbf{Mean fold macro OvR AUROC}} \\
\midrule
PS1-K3 & 0.939 \\
PS2-K3 & 0.942 \\
PS3-K3 & 0.954 \\
PS4-K3 & 0.936 \\
FS1-K3 & 0.933 \\
FS2-K3 & 0.945 \\
FSB-K3-18 & 0.898 \\
PS+-K3 & 0.941 \\
\bottomrule
\end{tabular}
}
\\[2pt]
{\scriptsize Methods adapt the corresponding binary recipes. FSB-K3-18 fits TRAIN PCA to 18 values (six experts' three-logit width) before FS1-K3; this does not match total capacity. PS+-K3 selects on SELECT among calibrated probability averaging and raw/richer prediction maps with logistic and fixed-XGBoost heads.}
\end{table}

SELECT chooses PS3-K3 in every fold: 0.954 versus FS1-K3's 0.933 and FS2-K3's 0.945; each FS method wins one fold. Xception and RECCE heads reach the iteration cap in every inner/final fit; RFM reaches it in multiple fits, with one final fit stopping at 498. Convergence is not established for all heads. These are task-specific heads on frozen binary-trained backbones, not end-to-end multiclass experts.

\textit{Fixed binary outputs.} Construction A uses three DF40-trained experts (CLIP-large, CLIP-base, Xception) to distinguish entire-face synthesis, face swap and reenactment: 20,633 rows in five folds. Construction B uses arch6 for GAN synthesis, diffusion and reenactment: 14,121 rows in three folds. FS-LR(std)-K3 is class-balanced multinomial logistic regression on TRAIN-standardized features. SELECT chooses PS among logistic/XGBoost heads on original logits and the richer maps of \app{B}. Each map searches 50/100/200 trees at depths 2/3/4; the selected trees reach both search maxima in every fold. Since all task rows are fake, \psmethod{H} temperatures use binary SELECT with real images and each fold's held-out content excluded. Class/fold manifests are released.

\begin{table}[ht]
\caption{Three-class classification from fixed binary expert outputs, using five (A) or three (B) outer folds holding out groups of generator families.}
\label{tab:k3_views}
\centering
{\footnotesize
\begin{tabular}{lrrr}
\toprule
\textbf{Construction} & \multicolumn{1}{l}{\begin{tabular}[b]{@{}l@{}}\textbf{Mean FS-LR(std)-K3}\\ \textbf{$-$ selected PS}\end{tabular}} & \multicolumn{1}{l}{\textbf{Positive folds}} & \multicolumn{1}{l}{\begin{tabular}[b]{@{}l@{}}\textbf{Minimum attainable}\\ \textbf{two-sided sign-flip $p$}\end{tabular}} \\
\midrule
A: entire-face synthesis, face swap, reenactment & +0.210 & 5/5 & 0.0625 \\
B: GAN synthesis, diffusion, reenactment & +0.201 & 3/3 & 0.2500 \\
\bottomrule
\end{tabular}
}
\\[2pt]
{\scriptsize Macro OvR AUROC equally weights classes within folds, then folds. Every selected XGBoost reaches 200 trees/depth 4, the grid boundaries. The final column is test resolution, not a significance claim.}
\end{table}

The constructions share examples and are not independent replications. All retained examples of A occur in its experts' upstream training partition under the path audit. B uses the same examples and outer folds in both output settings, but changes the FS preprocessing and PS search: the fixed-binary search omits the neural PS3-K3 selected with three-class heads. These few-fold comparisons therefore suggest task-aligned outputs can reduce feature benefit without isolating a causal effect of output width or establishing statistical sufficiency.

\paragraph{I.2. Synthetic speech.} We compare two datasets using 20 frozen experts: 16 binary heads on four speech encoders and four existing detectors. These checks are exploratory.

\textit{Experts and preprocessing.} Encoders are XLS-R 300M~\cite{babu2022xlsr} (\texttt{facebook/wav2vec2-xls-r-300m}), WavLM Large~\cite{chen2022wavlm} (\texttt{microsoft/wavlm-large}), HuBERT Large~\cite{hsu2021hubert} (\texttt{facebook/hubert-large-ll60k}) and Whisper medium~\cite{radford2023whisper} (\texttt{openai/whisper-medium}). We mean-pool layer-9 features from the first 4\,s at 16\,kHz, repeating short utterances. Layer 9 was selected among 6/9/12 by \fsmethod{1} macro AUROC on SpeechFake evaluation families, so selection uses evaluation data.

Each encoder has heads trained on text-to-speech, voice conversion, neural vocoding or all three. Heads use a 256-unit ReLU hidden layer, class-weighted cross-entropy, Adam (learning rate $10^{-3}$, weight decay $10^{-4}$), batch 256, 20 epochs and seed 1024. Other experts are AASIST/AASIST-L~\cite{jung2022aasist} with official weights\footnote{\url{https://github.com/clovaai/aasist}.}, RawNet2~\cite{tak2021rawnet2} with ASVspoof 2021 DF baseline weights\footnote{\url{https://github.com/asvspoof-challenge/2021/tree/main/DF/Baseline-RawNet2}.}, and a cross-speaker AASIST. The first three were trained on ASVspoof 2019 LA; training overlap for the fourth was not checked. They process 64,600 samples. Each expert supplies two logits; head hidden activations and detector embeddings give 5,600 feature values.

\textit{SpeechFake.} A bilingual 110,400-utterance subset~\cite{huang2025speechfake} uses genuine speech from LibriTTS, VCTK, AISHELL-1/3. A joint registry of 866 real/synthetic speakers separates head training (40,000), FIT (24,000), SELECT (8,000) and evaluation (38,400). HEAD, FIT and SELECT together contain 18 generator sources from 21 configured candidates; three (GlowTTS, DiffGANTTS and ProDiffTTS) contribute no sampled rows. Not every source appears in every partition. Evaluation holds out StyleTTS2, ParlerTTS, Tacotron2, FireRedTTS, SeedVC, FishSpeech, BigVGAN and WaveGlow. Each family has 2,500 genuine and 2,500 fake utterances, except FishSpeech (1,700 each).

\textit{Partition construction.} With seed 1024, held-out methods claim speaker IDs first, ordered by speaker scarcity; remaining speakers are allocated round-robin to HEAD, FIT and SELECT. Sampling balances genuine and synthetic utterances, with approximately equal TTS/VC/vocoder quotas. In HEAD, FIT and SELECT, each source is first sampled up to its share of the family quota, rounded up and bounded by the remaining quota. Remaining rows are sampled in source order. Only locally available files are eligible. Genuine utterances follow the joint speaker registry, with at most 80 genuine utterances per speaker per partition. Each family-specific head uses all genuine HEAD rows and that family's synthetic rows; the all-family head uses all HEAD rows. Speaker IDs and file paths are disjoint across partitions; transcript/prompt disjointness and language balance are not enforced.

\fsmethod{1}, \psmethod{1}, \psmethod{2}, \psmethod{A} and \psmethod{4} use Appendix A's settings. Trainable combiners fit FIT; temperature calibration uses SELECT. Evaluation macro AUROC retrospectively selects \psmethod{2}. The FS--PS difference is $-0.0007$ $[-0.0033,+0.0020]$; both exceed or equal 0.99 on seven families, while BigVGAN gives PS 0.705 and FS 0.700. A paired hierarchical percentile bootstrap uses 10,000 resamples (seed 1024), resampling families and then utterances. It fixes the fitted models and comparator and does not account for evaluation-based layer/comparator selection; it does not establish equivalence.

\textit{PartialSpoof.} We retain the same experts on PartialSpoof v1.2~\cite{zhang2023partialspoof}, refitting combiners only. With seed 1024, FIT uses all 2,580 genuine training utterances and 5,420 spoofed; SELECT takes 2,500 per class from development; evaluation uses all 7,355 genuine evaluation utterances and 17,645 spoofed. Cropping to roughly 4\,s may remove the spoofed segment while retaining its label; its frequency and effect were not measured.

FS uses FIT-standardized features, class-balanced logistic regression ($C=1$, L-BFGS, maximum 1,000 iterations). SELECT chooses \psmethod{2} among \psmethod{1}, \psmethod{2} and \psmethod{4}. Pooled evaluation AUROCs are 0.987 for FS and 0.974 for PS: +0.0133 $[+0.0114,+0.0153]$. The paired percentile interval uses 10,000 utterance resamples, seed 1024, without accounting for shared speakers or generators.

\begin{table}[!htbp]
\caption{Synthetic-speech comparisons on fixed partitions: speaker-disjoint SpeechFake partitions with held-out generator families, and PartialSpoof training/development/evaluation partitions. Combiners are not refitted per evaluation family.}
\label{tab:speech}
\centering
{\footnotesize
\setlength{\tabcolsep}{4pt}
\begin{tabular}{lrrrll}
\toprule
\textbf{Setting} & \multicolumn{1}{l}{\textbf{PS}} & \multicolumn{1}{l}{\textbf{FS}} & \multicolumn{1}{l}{\textbf{FS$-$PS}} & \textbf{95\% CI} & \textbf{Bootstrap unit} \\
\midrule
SpeechFake: eight held-out families & 0.961 & 0.960 & $-$0.0007 & $[-0.0033, +0.0020]$ & Families, then utterances \\
PartialSpoof: 25,000 utterances & 0.974 & 0.987 & +0.0133 & $[+0.0114, +0.0153]$ & Utterances \\
\bottomrule
\end{tabular}
}
\\[2pt]
{\scriptsize SpeechFake uses eight-family macro AUROC; PartialSpoof uses pooled utterance AUROC. FS preprocessing, PS selection and resampling differ as specified in the text; differences use unrounded values.}
\end{table}

Dataset, aggregation, FS preprocessing and PS selection differ between these studies, so the gap contrast does not isolate a cause or independently validate the vision findings. Removing all four existing detectors gives PS/FS 0.962/0.962 on SpeechFake and 0.971/0.986 on PartialSpoof under the same respective protocols. Public corpora, encoders and three detectors, together with the partition rules above, support approximate reimplementation. Local file availability can change the sampled subset.

\textit{Speech compute.} Recorded extraction estimates are approximately 15\,h for SpeechFake and 4\,h for PartialSpoof on an NVIDIA RTX 2070 Super with 8\,GB VRAM. Reanalysis from existing caches took 22.2\,min on a laptop CPU, including layer comparison, both composition analyses and their uncertainty calculations; head fitting and feature extraction were excluded. These are partial resource records, not total project costs: peak host memory, all head-training costs and additional exploratory or failed runs were not fully recorded.

\textit{Asset terms.} The SpeechFake paper specifies CC BY-NC 4.0 and component-specific terms. PartialSpoof v1.2 uses CC BY 4.0. XLS-R, HuBERT and Whisper use Apache-2.0, and WavLM uses CC BY-SA 3.0. The AASIST repository uses MIT. The ASVspoof 2021 DF baseline weights for RawNet2 have no stated license.

\paragraph{I.3. Cost and access.} With local checkpoints, logits and features share an expert forward pass; prediction-only APIs may not expose features. On arch6, 9,984 float16 feature values versus 12 float32 logits require 19,968 versus 48 bytes per sample before overhead. Their caches occupy 0.95\,GiB and 2.3\,MiB.

\begin{table}[!htbp]
\caption{Cost and access ledger for arch6. Timings and the TRAIN-matrix estimate use the standard split.}
\label{tab:cost}
\centering
{\footnotesize
\setlength{\tabcolsep}{4pt}
\begin{tabular}{llll}
\toprule
\textbf{Item} & \textbf{Feature interface} & \textbf{Prediction interface} & \textbf{Status} \\
\midrule
Values per sample & 9,984 float16; 19,968 B & 12 float32; 48 B & Measured \\
arch6 cache & 0.95 GiB & 2.3 MiB & Measured \\
Full 21-expert cache & 3.22 GiB & 8.2 MiB & Measured \\
Batch prediction time per sample & 35\,\textmu s (\fsmethod{1}) & 0.1\,\textmu s (\psmethod{2}) & Measured \\
Combiner fit and predict & 53 s (\fsmethod{1}) & 62 s (\psmethod{4}, including CV); 0.16 s (\psmethod{2}) & Measured \\
Float64 TRAIN matrix for \fsmethod{1} & 1.98 GiB & - & Estimated \\
\bottomrule
\end{tabular}
}
\\[2pt]
{\scriptsize arch6 array sizes sum standard-split TRAIN and SELECT plus the eight test sets, excluding labels, manifests and model weights. Timings are machine-dependent single measurements.}
\end{table}

Timings use already-loaded inputs, excluding expert inference/loading. Prediction time divides a 17,925-sample batch time by its size, not single-request latency; fitting uses 26,591 rows. Measurements are single runs (Windows, Python 3.11.3; processor/thread settings unrecorded). \psmethod{4} includes five-fold CV, while FS uses fixed settings, so tuning budgets differ. The 1.98\,GiB matrix estimate is not peak memory. Extraction throughput, transfer overhead and peak memory were not measured; these figures do not establish end-to-end deployment cost.

\subsection*{J. Evaluation Protocol Card}\label{sec:protocol-card}\label{app:J}

This card summarizes the four components of \S\ref{sec:protocol} for reuse, following the Data/Model Card format~\cite{gebru2021datasheets,mitchell2019modelcards}.

\paragraph{J.1. Intended regime.} Compare prediction- and feature-space combiners using the same frozen, independently trained experts and matched partitions. Experts expose predictions and penultimate features; only combiners train. Evidence is mainly binary, with limited three-class checks on binary-trained backbones (\app{I}.1).

\paragraph{J.2. Boundaries.} Joint expert optimization, selective routing, and many-class or structured-output tasks are untested. A dense gate using only predictions belongs to PS, one reading features to FS, and one reading raw inputs lies outside both; no such gate was evaluated. Prediction-only services may preclude FS access. Output dimensionality alone does not establish task sufficiency.

\paragraph{J.3. Screening and full-protocol checks.} Table~\ref{tab:tiers} distinguishes progressively broader checks. Its Minimum and Recommended columns are preliminary screening, not implementations of the full protocol. Section~\ref{sec:protocol} requires aggregation, linear stacking, neural stacking and a tuned nonlinear PS combiner, together with a linear FS anchor and a higher-capacity check, axis-isolated pools and grouped evaluation controls. The Stronger audit column includes these requirements and additional sensitivity checks. Specify expert interfaces, TRAIN/SELECT/TEST partitions and the held-out group. At every tier, inspect identifier/content overlap and expert-training provenance; address identified combiner train/test reuse in the primary comparison and report unresolved exposure. Grouped retraining cannot undo expert-training exposure. The tiers describe coverage of checks, not a guarantee of validity.

\begin{table}[ht]
\caption{Coverage of preliminary screening and extended audit checks.}
\label{tab:tiers}
\centering
{\footnotesize
\setlength{\tabcolsep}{4pt}
\begin{tabular}{>{\raggedright\arraybackslash}p{2.8cm}>{\raggedright\arraybackslash}p{2.9cm}>{\raggedright\arraybackslash}p{3.0cm}>{\raggedright\arraybackslash}p{4.0cm}}
\toprule
\textbf{Component} & \textbf{Minimum} & \textbf{Recommended} & \textbf{Stronger audit} \\
\midrule
(i) Baseline strength & Probability averaging and a tuned nonlinear stacker, with tuning confined to training data; this tier can overstate a feature gain (\app{F}) & Add linear stacking on logits & Add neural stacking, calibration and regularization variants, and richer functions of predictions; document search budgets \\
\midrule
(ii) Combiner capacity & A simple linear anchor, such as \fsmethod{1} on concatenated raw features & Add a higher-capacity combiner, such as \fsmethod{2} at one projection dimension & Sweep projection dimensions and compare the same combiner family across interfaces; add matched-input-width controls when making width-related claims \\
\midrule
(iii) Pool construction & One pool varying a single heterogeneity axis, with the other documented axes held fixed & Include architecture-diverse and training-data-diverse pools & Add boundary pools, such as weak-expert or low-heterogeneity pools, to examine the limits of the finding \\
\midrule
(iv) Overlap control and reporting granularity & Report overlap checks and exclusions, per-group results, and paired uncertainty at the held-out group level & Add grouped refitting where group transfer is the target, and assess the sensitivity of inference when groups are few & Examine repeated-content handling, alternative group weighting and complementary inferential summaries \\
\bottomrule
\end{tabular}
}
\\[2pt]
{\scriptsize Each tier extends the preceding one. DF40 method names illustrate choices; alternatives require documented settings and selection rules.}
\end{table}

\paragraph{J.4. Reporting.} For each pool and split, report:
\begin{enumerate}
\item Method-level macro AUROC, per-group FS/PS scores, class counts and positive/negative/tied differences; label secondary metrics and weighting choices.
\item PS candidates, search budgets and selection data; distinguish one comparator per pool from per-fold selection. TEST-selected ceilings are retrospective, with comparator-conditional intervals; deployment selection uses TRAIN/SELECT.
\item Identify the primary comparison and report a paired 95\% confidence interval at the held-out-group level, with the resampling scheme and seed. Use at least 10,000 bootstrap resamples; with few groups, add suitable inferential checks and state their assumptions and fold dependence. More resamples cannot compensate for few groups (\app{F}).
\item Input dimensions, per-model parameter counts, seed count and aggregation, and tuning budgets. Separate expert inference, storage/transfer and combiner costs when discussing deployment.
\end{enumerate}

\paragraph{J.5. Sensitivity.} Accompany each main conclusion, positive, negative or inconclusive, with a targeted check of baseline strength, overlap, reporting granularity or pool construction. State the choice changed, remaining untested axes and effects on magnitude, direction or scope. Capacity checks address an additional mechanism question. Identify exploratory analyses and contrasts that also change training composition, weights or comparator selection; report performed checks without treating confounded contrasts as causal effects.

\paragraph{J.6. Diagnostic risks.} Larger combiners may not help; high PS performance does not prove feature redundancy; poor transfer may affect both interfaces; means can hide uneven group support; and overlap or repeated content can affect PS and FS differently. Inspect these possibilities before generalizing an interface advantage.

\paragraph{J.7. Practical use.} Follow Section~\ref{sec:discussion}'s checks: establish strong prediction baselines, assess task-aligned outputs and expert transfer, price feature access, then run a held-out-group pilot. DF40 observations motivate these checks but supply no universal decision thresholds; representation geometry is descriptive.

\paragraph{J.8. Implementation.} Record checkpoints, preprocessing, layers and class order; freeze experts and use evaluation mode. Align interfaces by sample identifier/label. Keep all copies of shared content in a single partition and a single inner cross-validation fold, or remove duplicates; disclose residual overlap. Fit preprocessing/selection on the appropriate training data. Save seeds, settings, environment, predictions, group assignments and exclusions, distinguishing recomputed results from retained references.

\paragraph{J.9. Release.} The updated dataset card corrects its earlier claim that test generators were excluded from combiner fitting; \app{A} documents the standard-split overlap, and the cached arrays are unchanged.

 Code, stored results and instructions are at \anonrepo. The release provides reproduction harnesses for the original analyses and camera-ready additions, with commands in the README. The release manifest maps analyses to commands, outputs and tolerances. Verification-only mode compares stored values; the smoke test refits standard-split arch6 \fsmethod{1}; full mode refits supported DF40 analyses from the cache. Feature extraction is excluded. Missing upstream inputs, memory limits or disabled optional reruns can retain stored outputs; run records identify these cases. Separate scripts reproduce Appendix A's sensitivity checks.

The approximately 3.2\,GiB derived cache is at \datasethost, retaining its review address. It contains features, logits, labels, identifiers and split provenance, with checksums, a dataset card and Croissant 1.0 responsible-AI/provenance metadata. Code and our derived-cache contributions use CC BY-NC-SA 4.0; upstream DF40 and other component terms remain applicable. No raw media, detector weights or upstream DF40/DeepfakeBench code are bundled; component notices are in \texttt{camera\_ready/LICENSES.md}.

Per-sample predictions for the reported standard-split runs require recomputation; LOGO records include predictions and exclusions. GPU/software differences can change results, including deterministic fits. Normal deterministic checks use $10^{-4}$, fixed-output three-class/strict-refit checks $10^{-3}$, and GPU-trained three-class heads 0.005; installation documentation distinguishes the smoke tolerance. Only recomputed steps demonstrate reproduction in a new environment.

Figure~\ref{fig:tsne} can be regenerated from the released cache using a standalone plotting command. The main harness regenerates Figure~\ref{fig:bottleneck_scatter} from the full-ladder results (step C23). The original small-sample simulation script/output were not retained; the release rebuilds its detection-threshold analysis from the description. The release documents detailed execution coverage, resource estimates and platform variation.

\typeout{get arXiv to do 4 passes: Label(s) may have changed. Rerun}
\end{document}